\pdfoutput=1

\documentclass[11pt]{article}

\usepackage[final]{acl}

\usepackage{times}
\usepackage{latexsym}

\usepackage[T1]{fontenc}

\usepackage[utf8]{inputenc}

\usepackage{microtype}

\usepackage{inconsolata}

\usepackage{graphicx}

\usepackage[cjk]{kotex}
\usepackage[normalem]{ulem} %
\usepackage{amsmath}
\usepackage{booktabs} %
\usepackage{CJKutf8}
\usepackage{cleveref}
\usepackage{fdsymbol}
\usepackage{xspace}
\usepackage{pifont}
\usepackage{multirow}
\usepackage{rotating}
\usepackage[normalem]{ulem}
\useunder{\uline}{\ul}{}
\usepackage{svg}
\usepackage{float}
\usepackage[perpage]{footmisc}
\usepackage{tabularx}
\usepackage{array}

\graphicspath{ {./images/} }
\crefformat{section}{\S#2#1#3}
\crefformat{subsection}{\S#2#1#3}
\crefformat{subsubsection}{\S#2#1#3}

\useunder{\uline}{\ul}{} %
\def\eg{\emph{e.g}.,\xspace}

\newcommand{\cmark}{\textcolor[HTML]{00AA00}{\ding{51}}}%
\newcommand{\xmark}{\textcolor[HTML]{CC0000}{\ding{55}}}%

\newcommand{\mugemoji}{\raisebox{-1.5pt}{\includegraphics[width=0.9em]{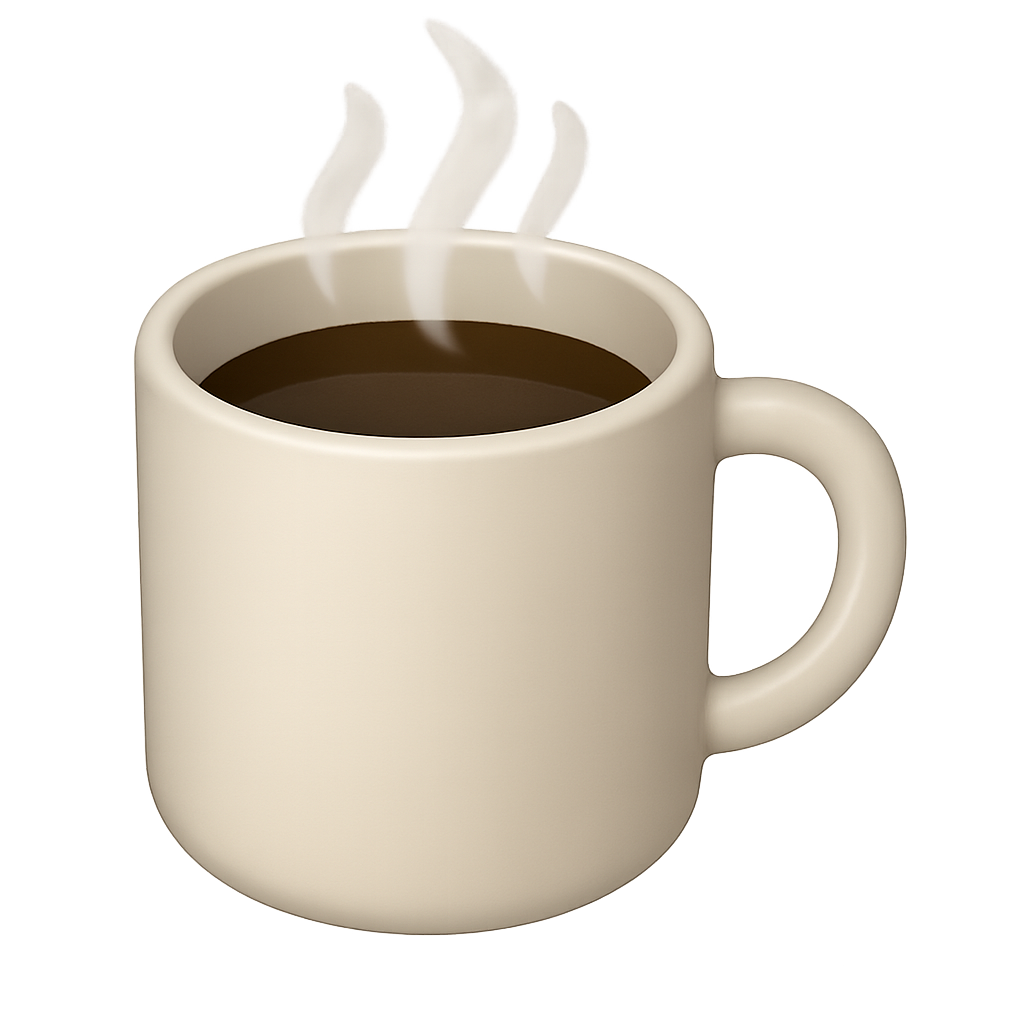}}\hspace{0.1em}}

\newcommand{\tool}{\textsc{MuG}-Eval\xspace}

\title{\mugemoji\tool: A Proxy Evaluation Framework \\ for Multilingual Generation Capabilities in Any Language}

\author{
  Seyoung Song$^{\heartsuit}$\thanks{These authors contributed equally.} \quad
  Seogyeong Jeong$^{\heartsuit}$\footnotemark[1] \quad
  Eunsu Kim$^{\heartsuit}$ \quad
  Jiho Jin$^{\heartsuit}$ \quad
  Dongkwan Kim$^{\heartsuit}$ \\
  \textbf{Jamin Shin}$^{\clubsuit}$ \quad
  \textbf{Alice Oh}$^\heartsuit$
  \\
  \\
  $^\heartsuit$KAIST \quad
  $^\clubsuit$Trillion Labs
  \\
  \small{\texttt{\{\href{mailto:seyoung.song@kaist.ac.kr}{\color{black}{seyoung.song}}, \href{mailto:sg.jeong28@kaist.ac.kr}{\color{black}{sg.jeong28}}, \href{mailto:kes0317@kaist.ac.kr}{\color{black}{kes0317}}, \href{mailto:jinjh0123@kaist.ac.kr}{\color{black}{jinjh0123}}, \href{mailto:dongkwan.kim@kaist.ac.kr}{\color{black}{dongkwan.kim}}\}@kaist.ac.kr}}
  \\
  \small{\texttt{\color{black}{jay@trillionlabs.co}, \color{black}{alice.oh@kaist.edu}}}
}

\begin{document}
\maketitle
\begin{abstract}
  Evaluating text generation capabilities of large language models (LLMs) is challenging, particularly for low-resource languages where methods for direct assessment are scarce.
We propose \mugemoji\tool, a novel framework that evaluates LLMs' multilingual generation capabilities by transforming existing benchmarks into conversational tasks and measuring the LLMs' accuracies on those tasks. We specifically designed these conversational tasks to require effective communication in the target language. Then, we simply use task success rate as a proxy for successful conversation generation.
Our approach offers two key advantages: it is independent of language-specific NLP tools or annotated datasets, which are limited for most languages, and it does not rely on LLMs-as-judges, whose evaluation quality degrades outside a few high-resource languages.
We evaluate 8 LLMs across 30 languages spanning high, mid, and low-resource categories, and we find that \tool correlates strongly with established benchmarks ($r$ > 0.75) while enabling standardized comparisons across languages and models.
Our framework provides a robust and resource-efficient solution for evaluating multilingual generation that can be extended to thousands of languages.

\end{abstract}

\section{Introduction}
\label{sec:introduction}
Large language models (LLMs) have demonstrated remarkable capabilities in many languages, but evaluating their multilingual generation abilities remains a significant challenge, particularly for low-resource languages.
These challenges are particularly pronounced for low-resource languages, which often lack robust natural language processing tools, comprehensive reference corpora, or established benchmarks.
Consequently, evaluation resources for these low-resource languages predominantly derive from massively multilingual evaluation benchmarks~\citep[\textit{inter alia}]{hasan-etal-2021-xl, goyal-etal-2022-flores, bandarkar-etal-2024-belebele, adelani-etal-2024-sib}.
Extending and evaluating natural language generation tasks presents considerable complexity, especially in the absence of language-specific resources.

\begin{figure}[t]
  \centering
  \includegraphics[width=\linewidth]{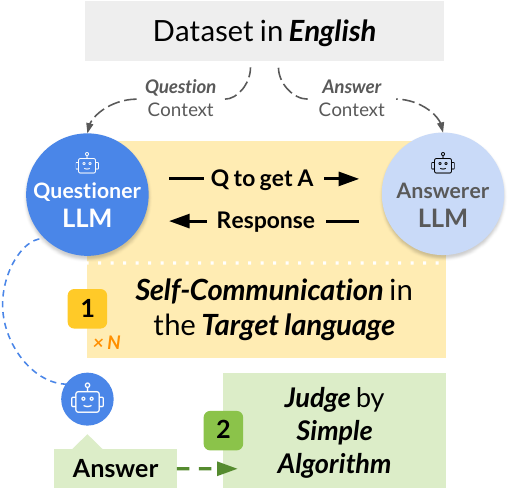}
  \caption{General concept of \mugemoji\tool. Two instances of the same LLM engage in self-communication in the target language to complete information-gap tasks. Model outputs are evaluated using algorithmic methods (\eg string matching or code testing), without requiring language-specific tools or LLMs-as-judges. Task success rate serves as a proxy for measuring the model's multilingual generation capability.}
  \label{fig:teaser}
\end{figure}

\begin{table*}[t]
  \centering
  \resizebox{\textwidth}{!}{%
    \begin{tabular}{@{}lcccccc@{}}
      \toprule
      Feature                                    & Global-MMLU & Belebele & Flores-101 & XL-Sum & MultiQ & \mugemoji{}\textbf{\tool}\\ \midrule
      Evaluates generation (not comprehension)  & \xmark & \xmark     & \cmark     & \cmark           & \cmark & \cmark                \\
      Metrics comparable across languages       & \cmark & \cmark     & \xmark     & \xmark           & \cmark & \cmark                 \\
      No LLMs-as-Judges required                & \cmark & \cmark     & \cmark     & \cmark           & \xmark   & \cmark                 \\
      Native speaker annotation is optional     & \xmark  & \xmark & \xmark & \xmark & \cmark & \cmark \\
      \# of languages supported                 & 42          & 122   & 101        & 47    & 137    & 2,102    \\ \bottomrule
    \end{tabular}%
  }
  \caption{Positioning of \tool among multilingual evaluation benchmarks. \tool uniquely combines: (1) evaluation of generation capability (not just comprehension), (2) cross-linguistically comparable metrics, and (3) objective scoring without LLMs-as-judges, and (4) reduced dependency on cross-lingual annotation. Tested on 30 languages, \tool currently supports 2,102 languages via GlotLID~\citep{kargaran-etal-2023-glotlid}, with the potential to scale further as more advanced language identification tools develop. Benchmarks referenced are MultiQ~\citep{holtermann-etal-2024-evaluating}, Flores-101~\citep{goyal-etal-2022-flores}, XL-Sum~\citep{hasan-etal-2021-xl}, Global-MMLU~\citep{singh2024global}, and Belebele~\citep{bandarkar-etal-2024-belebele}.}
  \label{tab:position}
\end{table*}

Recent approaches~\citep{holtermann-etal-2024-evaluating, pombal2025m} have employed LLMs-as-judges, but they face an inherent limitation---the reliability of judgments depends on the evaluator LLM's performance in the target language.
While this limitation may be less pronounced for high-resource languages~\citep{pombal2025m}, the applicability of such approaches to low-resource languages remains unclear and has not been rigorously validated.
Conventional evaluation approaches for generation ability often require human-annotated ground truth data, such as BLEU~\citep{papineni-etal-2002-bleu} for machine translation or ROUGE~\citep{lin-2004-rouge} for summarization.
Overall, there exists a gap in methodologies that offer both reliability and scalability for quantifying LLM generation performance across diverse languages.

In this paper, we propose \tool, a framework for evaluating the multilingual generation capabilities of LLMs, particularly for languages where direct evaluation proves challenging or infeasible.
Our methodology creates information-gap scenarios that require successful communication in the target language to complete tasks, such as providing hidden information to one agent while another must discover it through questioning.
We implement three tasks in \tool by adapting existing benchmarks into conversational and multilingual settings---Easy Twenty Questions~\citep{zhang-etal-2024-probing}, MCQ Conversation~\citep{bandarkar-etal-2024-belebele}, and Code Reconstruction~\citep{muennighoff2024octopack}---where task completion rates serve as proxies for different aspects of generation ability: reasoning, instruction following, and programming (\cref{sec:tasks}).
Our approach builds on the insight from~\citet{muennighoff2024octopack}: instead of directly assessing LLM-generated text quality, we can indirectly measure how well the LLM comprehends what it has itself generated.

We evaluate 8 LLMs across 30 languages from high-, mid-, and low-resource categories as defined by~\citet{singh-etal-2024-aya}.
Our experiments demonstrate that \tool has strong discriminative power, enabling precise comparisons both across languages and across models (\cref{sec:results}).
The framework shows high internal consistency among its three tasks and correlates strongly (Pearson's $r$ > 0.75) with established benchmarks including Belebele~\citep{bandarkar-etal-2024-belebele}, MultiQ~\citep{holtermann-etal-2024-evaluating}, and Global-MMLU~\citep{singh2024global} (\cref{sec:corr}).
Additionally, our analysis of MCQ Conversation reveals that when native-language references are unavailable, English is not always the optimal substitute language, particularly for low-resource languages (\cref{sec:dollar}).

Our primary contribution lies in proposing \tool\footnote{Code and dataset available at \url{https://github.com/seyoungsong/mugeval}.}, a novel language-agnostic framework for evaluating multilingual generation in large language models through self-comprehension tasks, without relying on language-specific NLP tools or human annotations.
To demonstrate the utility and effectiveness of this framework, we structure the paper as follows.
We begin by reviewing the landscape of multilingual generation evaluation, identifying critical gaps in existing methodologies that motivate our approach (\cref{sec:related_work}).
We then present the design of \tool, introducing three conversational tasks that recast generation evaluation as a communication-based task (\cref{sec:method}).
We evaluate eight large language models in 30 linguistically diverse languages, demonstrating strong correlations with established benchmarks while offering unprecedented scalability (\cref{sec:experiment}).
Through detailed analysis, we uncover cross-linguistic performance patterns and validate the effectiveness of \tool as a robust, language-agnostic evaluation framework (\cref{sec:discussion}),  and conclude with directions for future work in multilingual LLM evaluation (\cref{sec:conclusion}).

\section{Related Work}
\label{sec:related_work}

\begin{figure*}[t!]
  \centering  \includegraphics[width=\linewidth]{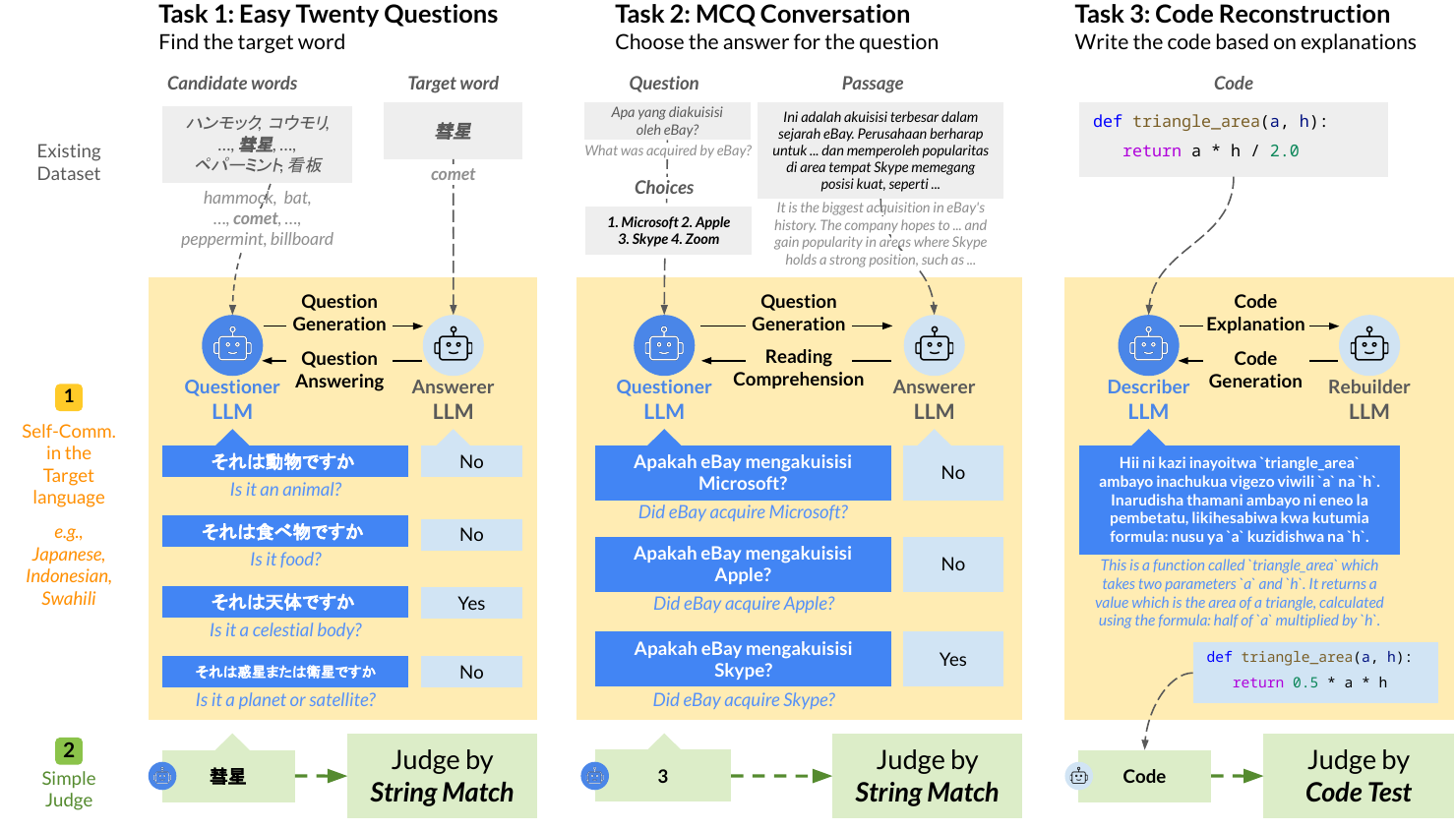}
  \caption{Overview of evaluation tasks. Two instances of the same LLM engage in self-communication in the target language to complete information-gap tasks: (1) Easy Twenty Questions---guessing a hidden word, (2) MCQ Conversation---finding the answer through passage-based dialogue, and (3) Code Reconstruction---explaining and reconstructing code.}
  \label{fig:tasks}
\end{figure*}

Reference-based metrics such as BLEU~\citep{papineni-etal-2002-bleu}, ROUGE~\citep{lin-2004-rouge}, and chrF~\citep{popovic-2015-chrf} assess generation quality by comparing outputs against reference texts, usually requiring human-generated target texts as ground truth.
These metrics are widely adopted in benchmarks such as MEGA~\citep{ahuja-etal-2023-mega}, GlotEval~\citep{luo2025gloteval}, Multi-IF~\citep{he2024multiifbenchmarkingllmsmultiturn}, and BenchMAX~\citep{huang2025benchmax}.
However, such reference-based approaches are limited by their reliance on high-quality parallel data, which is scarce in many languages.
Moreover, they struggle in cross-lingual comparisons due to their sensitivity to lexical and syntactic features.

To address these limitations, reference-free methods—particularly those using LLMs as evaluators—gained attention~\citep{dang2024ayaexpansecombiningresearch, holtermann-etal-2024-evaluating, pombal2025m}.
Nonetheless, \citet{hada-etal-2024-large} highlights the instability and reduced reliability of LLM evaluators in low-resource or non-Latin script languages, raising concerns about fairness and generalizability.

An emerging line of work evaluates generation quality through downstream utility, assessing how well generated content supports task completion.
Recent benchmarks explore the generation-comprehension link through interactive information-gap tasks that require mutual understanding.
These include clarifying question generation~\citep{gan2024clarq}, reference games~\citep{gul-artzi-2024-cogen, eisenstein2023md3}, bidirectional code understanding~\citep{muennighoff2024octopack}, and multi-turn interactive benchmarks such as HumanEvalComm~\citep{wu2024benchmarking}, telephone-game simulations~\citep{perez2025when}, and 20Q~\citep{zhang-etal-2024-probing}.

Drawing inspiration from 20Q~\citep{zhang-etal-2024-probing} and HumanEvalExplain~\citep{muennighoff2024octopack}, our framework builds on tasks that inherently require both comprehension and generation, foregrounding successful communication as the central evaluation criterion.
Designed to be language-agnostic, reference-free, and LLM-independent, it offers a more equitable and scalable multilingual evaluation across an unlimited spectrum of languages.

\section{\mugemoji{}\tool: A Language-Agnostic Evaluation Framework}
\label{sec:method}

\tool consists of three tasks adapted from existing benchmarks~\citep{zhang-etal-2024-probing, bandarkar-etal-2024-belebele, muennighoff2024octopack} to evaluate multilingual generation capabilities.
The benchmarks for Easy Twenty Questions and Code Reconstruction were originally English-only, while the source for the MCQ Conversation task is the multilingual Belebele dataset.
Each task is structured as a self-communication scenario between two ``LLM instances''---separate API calls to the same model, each assigned a distinct conversational role (\eg Questioner or Answerer) with a unique system prompt and access to different information.
The instances communicate turn-by-turn in the target language, with the output from one serving as the input for the next.
The model's capability is measured by the task completion rate, which serves as the primary evaluation metric.

This section provides detailed descriptions of each task and evaluation procedures.
Additional details, including prompts and generation parameters, are provided in the Appendix~\ref{sec:appx_generations}.

\subsection{Tasks}
\label{sec:tasks}

\paragraph{Easy Twenty Questions.}

This task evaluates reasoning and strategic questioning abilities through a word-guessing game.
Drawing from the \emph{Things} dataset~\citep{zhang-etal-2024-probing}, we translate 140 English words into 30 languages using Google Translate.
One model instance (answerer) receives a hidden word from this set, while another (questioner) must identify it from a list of 100 candidates.
The questioner poses up to 20 yes/no questions in the target language, to which the answerer responds only with ``yes,'' ``no,'' or ``maybe'' in English.
The predefined candidates ensure consistent evaluation across languages, mitigating lexical diversity from affecting task difficulty or scoring mechanisms.

\paragraph{MCQ Conversation.}

We transform the Belebele benchmark~\citep{bandarkar-etal-2024-belebele}—a reading comprehension dataset spanning 122 languages—into a conversational task.
From the original dataset of 900 samples, we separate the reading passages from their corresponding questions and answer choices.
Similar to the previous task, the answerer instance accesses only the passage, while the questioner sees the question and four answer options.
To discover the correct answer, the questioner may ask up to 10 yes/no questions in the target language, receiving ``yes,'' ``no,'' or ``maybe'' responses in English, similar to the previous task.
This design tests multi-turn instruction-following capabilities.

\paragraph{Code Reconstruction.}

This task adapts HumanEvalExplain~\citep{muennighoff2024octopack} to assess code generation abilities across languages, not only in English.
Using 164 Python function samples with corresponding unit tests, one model instance (describer) generates a natural language explanation of the code in the target language.
Another instance (rebuilder) then reconstructs the original function from this description and the function declaration snippet.
Success is measured by whether the reconstructed code passes all unit tests.

\subsection{Evaluation Metrics}

Task completion rate serves as our primary metric, calculated as the ratio of successfully completed tasks.
We use exact string matching for word or choice predictions, with responses prompted to appear within double brackets and extracted via regular expressions.
We employ GlotLID~\citep{kargaran-etal-2023-glotlid} to ensure the model's responses are in the target language.
Tasks fail when models: (1) produce a question or description in the wrong language, (2) produce invalid responses, or (3) violate task-specific constraints such as including more than 20 consecutive source code characters in explanations.

\section{Experiments}
\label{sec:experiment}

\begin{figure*}[t!]
  \centering
  \includegraphics[width=\linewidth]{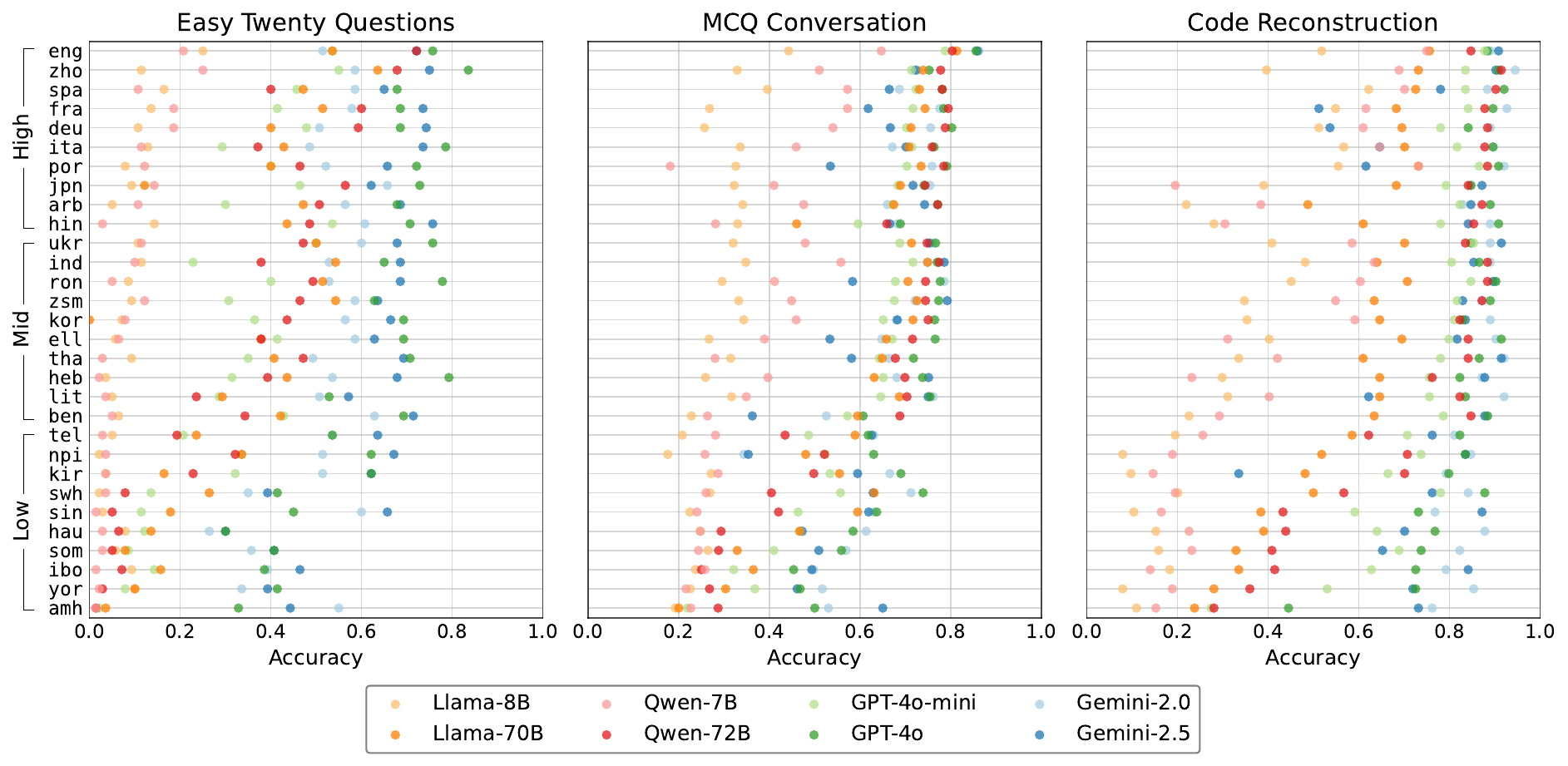}
  \caption{Accuracy of 8 LLMs across three tasks in 30 languages.  Languages are grouped by resource level and sorted by average performance within each group. Results show that Code Reconstruction is the easiest task, followed by MCQ Conversation and Easy Twenty Questions. The gap is minor between high and mid-resource languages, but substantial between mid and low. Larger models consistently outperform smaller ones within the same language family, and tasks exhibit distinct ceiling effect.}
  \label{fig:main_figure}
\end{figure*}

\begin{table*}[ht]
  \centering
  \resizebox{\textwidth}{!}{%
    \begin{tabular}{@{}l|rrrrr|rrrrr|rrrrr@{}}
      \toprule
      \multirow{2}{*}{Model} & \multicolumn{5}{c|}{Easy Twenty Questions}                                                                                    & \multicolumn{5}{c|}{MCQ Conversation}                                                                                            & \multicolumn{5}{c}{Code Reconstruction}                                                                                       \\
      & \multicolumn{1}{c}{All} & \multicolumn{1}{c}{\textsc{Eng}} & \multicolumn{1}{c}{High} & \multicolumn{1}{c}{Mid} & \multicolumn{1}{c|}{Low} & \multicolumn{1}{c}{All} & \multicolumn{1}{c}{\textsc{Eng}} & \multicolumn{1}{c}{High} & \multicolumn{1}{c}{Mid} & \multicolumn{1}{c|}{Low} & \multicolumn{1}{c}{All} & \multicolumn{1}{c}{\textsc{Eng}} & \multicolumn{1}{c}{High} & \multicolumn{1}{c}{Mid} & \multicolumn{1}{c}{Low} \\ \midrule
      GPT-4o                 & {\ul 62.21}             & \textbf{75.71}          & \textbf{72.64}         & \textbf{69.21}         & {\ul 44.79}             & \textbf{70.14}          & 85.56                   & \textbf{77.31}         & \textbf{74.33}         & \textbf{58.78}          & {\ul 83.43}             & 88.41                   & {\ul 89.02}            & {\ul 86.59}            & {\ul 74.70}            \\
      Gemini-2.0-flash       & 51.93                   & 51.43                   & 56.07                  & 55.57                  & 44.14                   & {\ul 66.72}             & \textbf{86.22}          & 73.33                  & 69.74                  & {\ul 57.08}             & \textbf{86.79}          & {\ul 89.02}             & \textbf{89.21}         & \textbf{89.45}         & \textbf{81.71}         \\
      Gemini-2.5-flash       & \textbf{62.26}          & {\ul 72.14}             & {\ul 70.57}            & {\ul 66.36}            & \textbf{49.86}          & 62.90                   & {\ul 85.89}             & 68.90                  & 65.74                  & 54.07                   & 77.05                   & \textbf{90.85}          & 74.63                  & 84.39                  & 72.13                  \\
      Qwen2.5-72B            & 35.17                   & {\ul 72.14}             & 53.86                  & 40.64                  & 11.00                   & 61.90                   & 80.33                   & {\ul 76.61}            & {\ul 72.44}            & 36.63                   & 73.68                   & 84.76                   & 87.56                  & 84.15                  & 49.33                  \\
      GPT-4o-mini            & 31.95                   & 53.57                   & 44.29                  & 35.93                  & 15.64                   & 59.83                   & 78.78                   & 70.11                  & 65.91                  & 43.48                   & 75.02                   & 87.80                   & 82.50                  & 80.12                  & 62.44                  \\
      Llama-3.3-70B          & 33.79                   & 53.57                   & 44.14                  & 40.36                  & 16.86                   & 61.15                   & 81.33                   & 70.04                  & 68.29                  & 45.12                   & 58.03                   & 75.61                   & 68.05                  & 65.61                  & 40.43                  \\
      Qwen2.5-7B             & 7.90                    & 20.71                   & 14.50                  & 6.64                   & 2.57                    & 37.33                   & 64.67                   & 46.48                  & 40.33                  & 25.17                   & 40.47                   & 75.00                   & 56.28                  & 46.22                  & 18.90                  \\
      Llama-3.1-8B           & 8.45                    & 25.00                   & 12.64                  & 7.71                   & 5.00                    & 28.94                   & 44.22                   & 33.46                  & 30.23                  & 23.13                   & 31.95                   & 51.83                   & 46.10                  & 36.16                  & 13.60                  \\ \bottomrule
    \end{tabular}%
  }
  \caption{Average accuracy (\%) of 8 LLMs across three tasks, grouped by language resource categories.
    The best and the second-best performances within each task and resource category are \textbf{bolded} and {\ul underlined}, respectively.
    A consistent performance degradation is observed as the language resource level decreases from high (including English) to low.
  }
  \label{tab:main_table}
\end{table*}

\paragraph{Models.}

We evaluate eight multilingual large language models to assess their generation capabilities across diverse languages.
Our selection includes four open-weight models: Llama 3.3-70B~\citep{grattafiori2024llama}, Llama 3.1-8B, Qwen2.5-72B~\citep{yang2024qwen2}, and Qwen2.5-7B, alongside four closed-source models: GPT-4o~\citep{gpt4o}, GPT-4o-mini, Gemini 2.5 Flash~\citep{gemini25}, and Gemini 2.0 Flash~\citep{gemini2}.
All models are accessed via API endpoints, with GPT-4o variants served through Azure OpenAI Services and the remaining models through OpenRouter.
Detailed model information is provided in the Appendix~\ref{sec:appx_models}.

\paragraph{Languages.}

We test our framework on 30 languages grouped by resource availability following~\citet{singh-etal-2024-aya}'s classification, with 10 languages selected from each resource category.
We include high-resource languages Arabic (\texttt{arb}), Chinese (\texttt{zho}), English (\texttt{eng}), French (\texttt{fra}), German (\texttt{deu}), Hindi (\texttt{hin}), Italian (\texttt{ita}), Japanese (\texttt{jpn}), Portuguese (\texttt{por}), and Spanish (\texttt{spa});
mid-resource languages  Bengali (\texttt{ben}), Greek (\texttt{ell}), Hebrew (\texttt{heb}), Indonesian (\texttt{ind}), Korean (\texttt{kor}), Lithuanian (\texttt{lit}), Malay (\texttt{zsm}), Romanian (\texttt{ron}), Thai (\texttt{tha}), and Ukrainian (\texttt{ukr}); and low-resource languages Amharic (\texttt{amh}), Hausa (\texttt{hau}), Igbo (\texttt{ibo}), Kyrgyz (\texttt{kir}), Nepali (\texttt{npi}), Sinhala (\texttt{sin}), Somali (\texttt{som}), Swahili (\texttt{swh}), Telugu (\texttt{tel}), and Yoruba (\texttt{yor}).
This selection covers diverse language families and writing systems, including Latin, Cyrillic, and Devanagari scripts, ensuring comprehensive evaluation across typologically distinct languages.
Detailed language information is provided in the Appendix~\ref{sec:appx_langs}.

\subsection{Results}
\label{sec:results}

Table~\ref{tab:main_table} summarizes overall accuracy, and Figure~\ref{fig:main_figure} visualizes trends by language and task. Full results are provided in Appendix~\ref{sec:appx_all_result}.

\paragraph{How difficult is \tool{}?}
Average accuracy scores across tasks vary depending on the model and the resource level of the language.
Code Reconstruction is the easiest task, followed by MCQ Conversation, while Easy Twenty Questions challenges the most.
This may be due to the number of interaction turns: multi-turn tasks are more error-prone as mistakes accumulate. This pattern aligns with average turn counts (Table~\ref{tab:stats}): Easy Twenty Questions requires the most turns, MCQ Conversation fewer, and Code Reconstruction only one.

\paragraph{Performance varies across resource levels and models.}
The performance gap between high- and mid-resource language groups is relatively small compared to the much larger gap observed between mid- and low-resource groups.
Additionally, larger models consistently outperform smaller ones within the same model family.
Despite some variation in task-wise rankings, overall trends of task rankings remain stable across models.

\paragraph{Complementary ceiling effects exist across tasks.}
Code Reconstruction and MCQ Conversation saturate near the upper bound—around 0.9 and 0.8, indicating 90\% and 80\% accuracy.
In contrast, Easy Twenty Questions exhibits saturation toward the lower end, with many scores concentrated near zero—especially in low-resource languages and smaller models.
MCQ Conversation shows lower saturation than its original benchmark, Belebele (0.8 vs. 0.95; see Figure~\ref{fig:violin}), likely due to its split-agent design, which can produce ambiguous question generations, leading to unsolvable cases.

These differing saturation patterns enhance the discriminative power of \tool.
Easier tasks are more effective at separating weaker models and low-resource languages, while the harder task better distinguishes stronger models and high-resource languages. Together, they ensure that \tool maintains discriminative power across the full performance spectrum.

\section{Discussion}
\label{sec:discussion}

\subsection{Comparative Analysis}
\label{sec:corr}

\paragraph{Which tasks best distinguish between models?}
\begin{figure}[ht!]
  \centering
  \includegraphics[width=\linewidth]{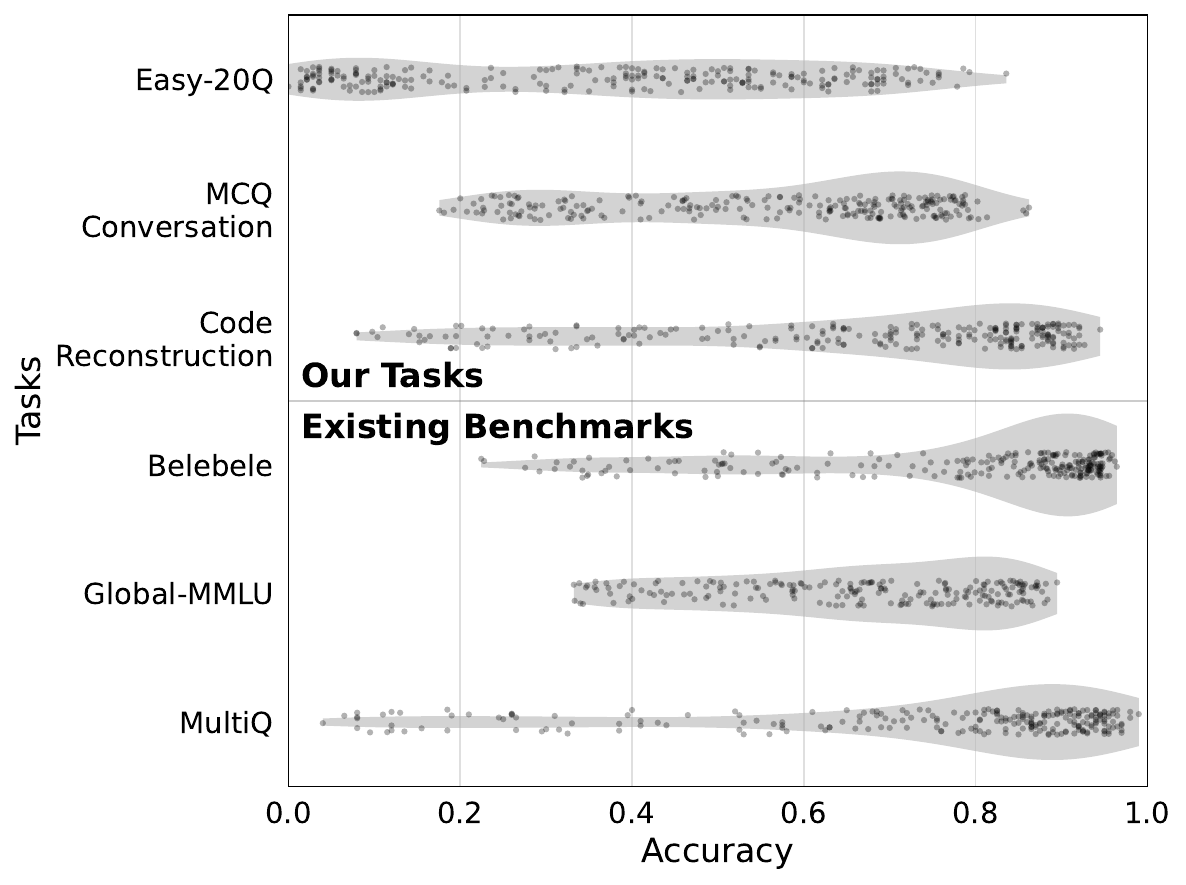}
  \caption{Score distributions across six evaluation tasks, demonstrating varying discriminative powers.
    Notably, MCQ Conversation, derived from the Belebele task, exhibits greater statistical dispersion, indicating greater ability to distinguish between models than the original Belebele benchmark.
  }
  \label{fig:violin}
\end{figure}

Figure~\ref{fig:violin} presents violin plots of accuracy scores for six tasks, including the three introduced in \tool.
Easy Twenty Questions exhibited a broad distribution of scores, indicating strong discriminative power and the ability to distinguish models with varying capabilities.
In contrast, Code Reconstruction showed a much narrower range, suggesting limited differentiation among a few models.
Notably, \tool's MCQ Conversation demonstrated substantially greater discriminative power compared to the original Belebele task, highlighting its usefulness in evaluating multilingual understanding with finer granularity. Overall, all three tasks in \tool show greater discriminative capability than the three existing benchmarks.

\paragraph{How consistent is performance across different tasks?}

\begin{figure}[ht]
  \centering
  \includegraphics[width=\linewidth]{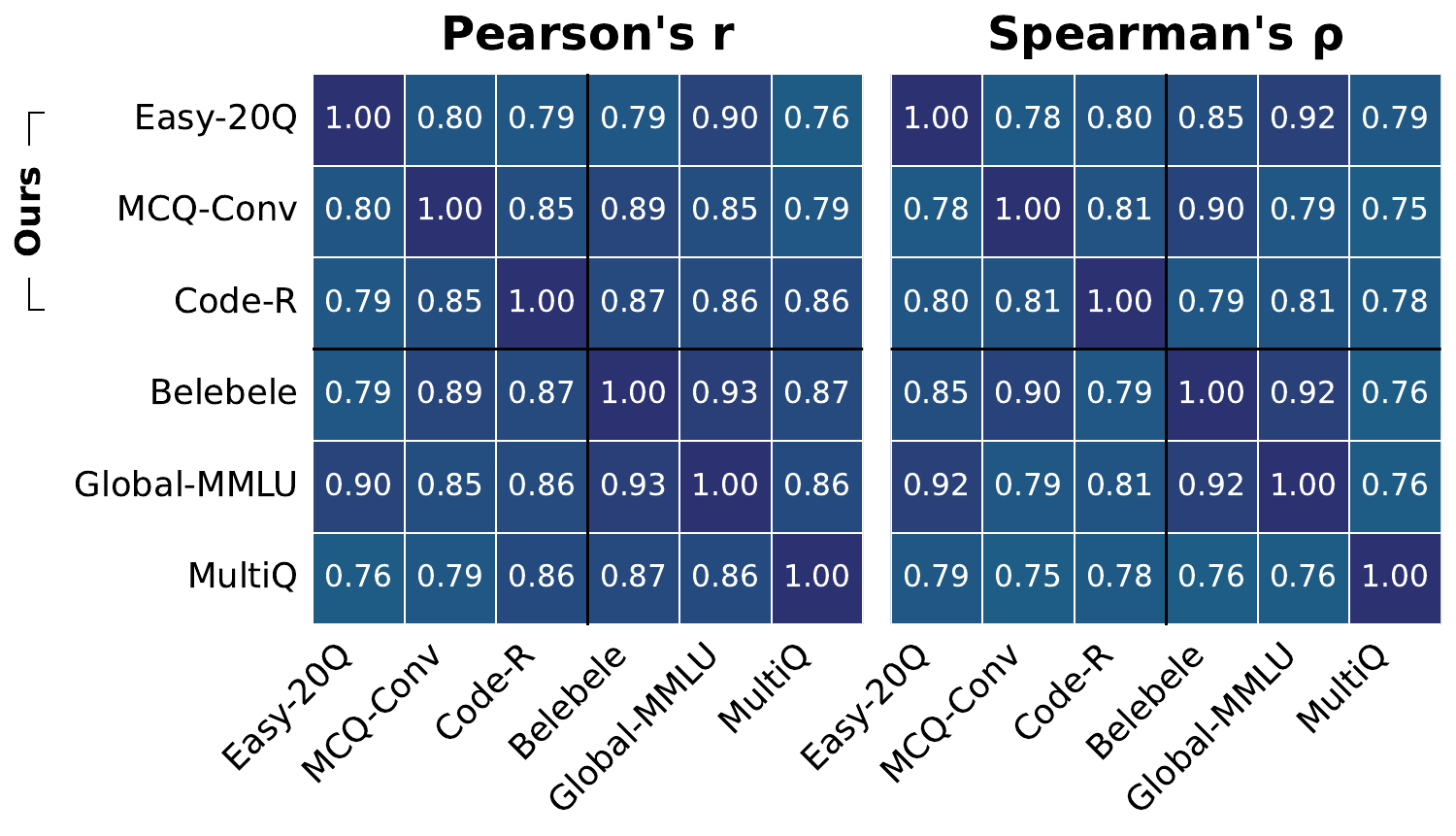}
  \caption{Correlation analysis between \tool{} tasks and existing multilingual benchmarks. Heatmaps show Pearson's $r$ (left) and Spearman's $\rho$ (right) correlation coefficients between three \tool{} tasks and three established benchmarks. All correlations exceed 0.75, demonstrating strong consistency between \tool{} and existing evaluation methods, validating its effectiveness as a multilingual evaluation framework.}
  \label{fig:comparision_benchmarks}
\end{figure}

To validate the internal consistency of our framework, we analyzed performance correlations across our three tasks.
While the tasks measure distinct abilities, a moderate positive correlation suggests that they capture a consistent, general signal of a model's multilingual capabilities.
Figure~\ref{fig:comparision_benchmarks} compares these performance correlations across six tasks, including the three introduced in \tool.
Pearson correlation coefficients are all above 0.75, indicating strong consistency between task accuracy. Spearman's rank correlation coefficients exceed 0.75 in all cases, suggesting positive correlations in rank ordering.
The reason why the correlations are not perfect is likely due to the distinct capabilities each task targets.
Easy Twenty Questions primarily evaluates the reasoning aspect of generation, MCQ Conversation focuses on instruction following, Code Reconstruction assesses coding under information asymmetry.
These differences account for the variation observed across tasks despite overall similarity.

\paragraph{Validation against established benchmarks.}

Figure~\ref{fig:comparision_benchmarks} also compares performance correlations across six tasks, including the three introduced in \tool.
While neither Pearson's nor Spearman's coefficients indicate perfect alignment between the three tasks in \tool and existing benchmarks, the figure demonstrates a high degree of correlation.
This suggests that \tool produces reliable results in terms of both accuracy and ranking, despite its low cost due to the absence of human-annotated datasets.
The detailed visualization result on Pearson's $r$ is provided in Appendix~\ref{appx:visual_6by6}.

\subsection{Language Resource Flexibility: A Substitution Analysis}
\label{sec:dollar}

\begin{figure}[ht]
  \centering
  \includegraphics[width=0.9\linewidth]{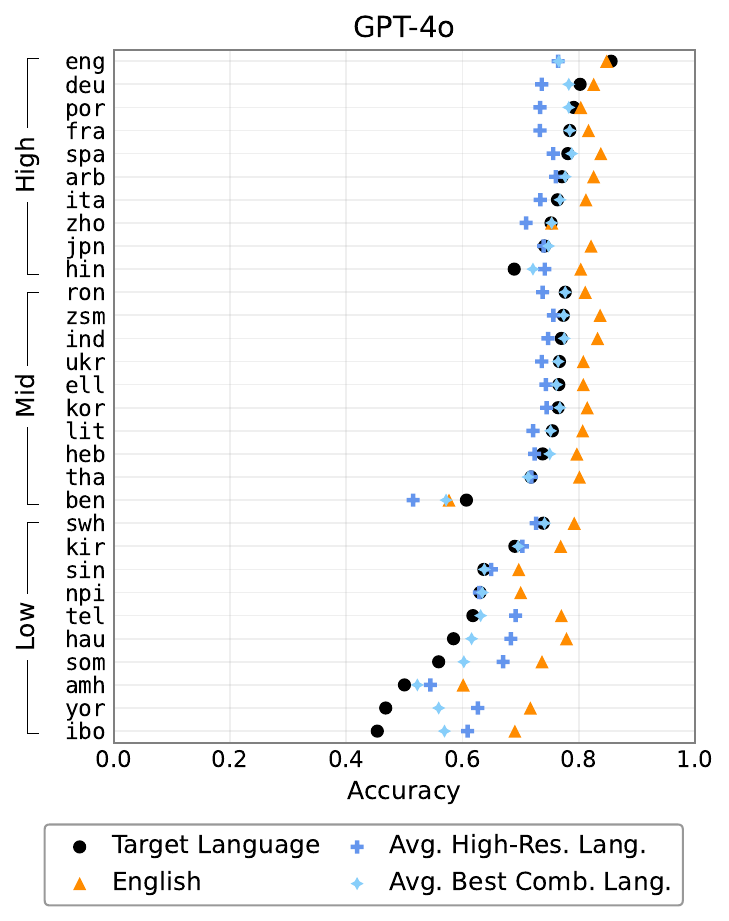}
  \caption{MCQ Conversation accuracy comparison across 30 languages for GPT-4o using passages in: (1) the target language, (2) English, and (3) five fixed high-resource languages (averaged), and (4)an optimized subset of up to five high-resource languages most similar to the target language. Results demonstrate that high-resource language substitution more closely approximates native language performance than using English alone, especially for low-resource languages.}
  \label{fig:dollar_index}
\end{figure}

The original MCQ Conversation task assumes that the answerer receives a passage written in the target language.
This raises a practical question: if such a passage is unavailable, can an English passage be used instead without significantly affecting performance?
Would using passages from other high-resource languages yield a better substitute?

To investigate this, three experimental settings were compared: (1) using the original target language passage, (2) using an English passage, and (3) using five separate versions of each passage, each written in one of the high-resource languages—English, Chinese, Arabic, Japanese, or Hindi.
Two models, GPT-4o and GPT-4o-mini, were evaluated,\footnote{This resource-intensive analysis was limited to the GPT models available via Azure OpenAI Service to stay within our computational budget.} with the GPT-4o result presented in Figure~\ref{fig:dollar_index}.
The result on the other model (GPT-4o-mini) is provided in the Appendix~\ref{sec:appx_mini_plot_result}.

On average, performance based on the five high-resource language passages more closely approximated that of the target-language baseline than when using English alone. This indicates that incorporating diverse high-resource languages may provide a better alternative when native-language passages are unavailable.

To further validate the applicability of MCQ Conversation, we conducted an evaluation to assess whether replacing native-language passages with those in five high-resource languages maintains consistent performance patterns across languages. The correlation between results using original target-language passages and those using the high-resource substitutes was 0.60 for Pearson (based on raw scores) and 0.71 for Spearman (based on rank-order consistency). Given that \tool is ultimately designed for cross-lingual comparisons, the higher Spearman correlation suggests that relative language rankings are preserved without native-language input.

To deepen the analysis, we identified the high-resource language combination that best approximates the native passage for each target language.
MCQ Conversation was executed across all target languages using the five high-resource passages across two models: GPT-4o and GPT-4o-mini.

For each case, the L2 distance between the performance with the substituted passage and that on the original native-language passage was calculated.
The combination of high-resource language that minimizes this distance is reported in Table~\ref{tab:dollar_lang} and plotted in Figure~\ref{fig:dollar_index}.
Results show that for high- and mid-resource languages, the best-performing combination typically includes English.
However, for low-resource languages, combinations excluding English usually performed better.
This indicates that English is not always the optimal substitute, especially for low-resource languages.
The details about the best combinations on each language is provided in Appendix~\ref{appx:appx_mcqconv_comb}.

\subsection{Qualitative Error Analysis: GPT-4o in English and Korean}
\label{sec:quali}

\begin{figure}[ht!]
  \centering
  \includegraphics[width=\linewidth]{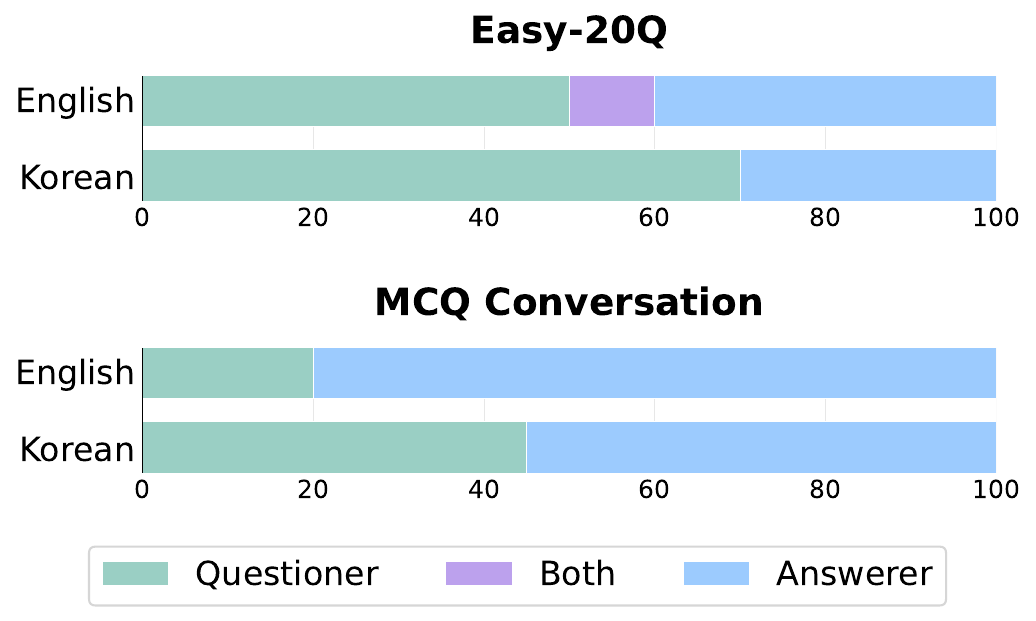}
  \caption{Attribution of errors by conversational role. Bars show the percentage of failures caused by Questioner (green), Answerer (blue), or Both roles (purple).}
  \label{fig:quali}
\end{figure}

\paragraph{Setup.}
To validate that task completion rates reflect genuine language capabilities, we conducted a fine-grained error analysis on GPT-4o outputs in English and Korean.
We chose GPT-4o as a representative high-performing model and selected English and Korean to leverage the authors' proficiency for reliable annotation.
The authors manually annotated 160 GPT-4o conversation logs, sampling 20 success and failure cases each for Easy Twenty Questions and MCQ Conversation in English and Korean.
Initial classification was performed using Gemini-2.5-flash, then manually corrected by two authors proficient in both languages.

\paragraph{Findings.}
Figure~\ref{fig:quali} reveals systematic task-specific error patterns that validate our framework design.
The Code Reconstruction task is excluded from this role-based error analysis, as attributing failure to either the `describer' or `rebuilder' is inherently ambiguous.
Easy Twenty Questions failed primarily due to questioner errors, reflecting strategic question generation challenges, while MCQ Conversation showed predominantly answerer errors, indicating passage comprehension difficulties.
These patterns remained consistent across languages, confirming that failures stem from genuine communicative challenges rather than external factors.
Success cases showed minimal errors in both roles, while rare successful cases with conversational errors reflected expected random chance.
The LLM-based initial annotation achieved 78.8\% accuracy (62.5\% for failure cases, 95.0\% for success cases).

\paragraph{Representative Error Case.}
In the MCQ Conversation task, Questioner errors often stemmed from failures to faithfully incorporate all relevant information from the original query when generating questions. Key semantic or lexical elements were frequently omitted, resulting in questions that lacked sufficient grounding in the passage—ultimately leading to unanswerable or misleading queries. In contrast, Answerer errors primarily reflected incorrect inference from the passage. Detailed examples of representative error cases are provided in Appendix~\ref{appx:quali}.

In the Easy Twenty Questions task, Questioner errors were typically caused by ineffective information-seeking strategies, such as asking insufficiently discriminative questions within the 20-turn limit or making premature guesses despite the presence of multiple plausible candidates. Most Answerer errors in this task were due to hallucinated responses, where the model generated logically incorrect ``yes''/``no''/``maybe'' answers.

\subsection{Generation Statistics}
\label{sec:stats}

While running the experiments, we collected detailed generation statistics, averaged over models and language groups. Specifically, we measured (1) token count, (2) sequence length, (3) language fidelity, (4) instruction-following of the Answerer, and (5) interaction length. A full description of these statistics is provided in Appendix~\ref{appdx:stats}. We summarize key findings below:
\begin{itemize}
  \item \textbf{Token Count and Sequence Length}: Output length varied by language resource level, with English being the shortest and low-resource languages generally producing the longest outputs.
  \item \textbf{Language Fidelity}: Although slightly lower in low-resource languages, fidelity scores remained similarly high across all groups.
  \item \textbf{Answerer Instruction-Following and Interaction Length}: These metrics were largely consistent across language resource groups and models. On average, Easy Twenty Questions involved 14.3 turns, and MCQ Conversation 4.0.
\end{itemize}

\section{Conclusion}
\label{sec:conclusion}

A fundamental limitation in multilingual evaluation is the reliance on ground-truth references or LLM-based judgments, which are often unreliable or infeasible for low-resource languages.
To address this, we introduce \mugemoji{}\textbf{\tool}, a language-agnostic evaluation framework based on three conversational task completion between LLMs that assess both generation and comprehension.

We evaluate 8 LLMs across 30 languages using \tool. Our framework demonstrates strong internal consistency and aligns well with established multilingual benchmarks, while remaining reference-free and cost-effective.
Our results highlight a few implications. First, \tool enables fine-grained performance comparisons even in low-resource settings due to its task diversity and saturation characteristics. Second, we find that substituting native-language passages with English often degrades performance—especially for low-resource languages—underscoring the need for evaluation methods that go beyond English-centric assumptions.

\section*{Limitations}
\label{sec:limitations}

\tool measures whether communication succeeds, but not how well it succeeds—a model generating minimal functional text scores identically to one producing sophisticated, nuanced output, as long as both complete the task.
This limitation poses challenges for applications requiring natural, culturally appropriate, or stylistically rich text generation.
Furthermore, comparing linguistic quality across languages remains fundamentally difficult because notions of richness and quality vary significantly across linguistic and cultural contexts, making it challenging to establish universal cross-linguistic metrics.
This focus on communicative effectiveness over stylistic quality is an intentional design choice, ensuring our framework remains scalable and objective in low-resource settings where fluency evaluation is often infeasible.
While this trade-off enables our language-agnostic evaluation approach, it remains a limitation for comprehensively assessing generation quality.

While \tool's reliability is supported by its strong correlations with existing benchmarks, comprehensive human evaluation has not yet been conducted.
Our qualitative error analysis of 160 conversation logs (\cref{sec:quali}) provided initial validation of failure patterns and confirmed that task failures stem from genuine communicative challenges rather than external factors.
However, broader human validation across all 30 languages would provide deeper insights into the framework's fairness across different languages and enable more detailed qualitative analysis of model performance patterns.
Given the conversational nature of \tool's tasks, human evaluation could reveal which specific conversational aspects challenge different models, particularly since performance varies significantly depending on conversational roles.

Despite \tool's language-agnostic design, certain implementation aspects remain English-centric.
The difficulty of accurately translating prompts into all target languages, especially low-resource ones, necessitated using English for instructional prompts in the conversational scenarios.
Additionally, the Code Reconstruction task employs Latin script for code, with variable and function names following English naming conventions.
These factors may introduce systematic biases against non-Latin script languages and low-resource language contexts, potentially affecting the framework's cross-linguistic validity.

\section*{Ethical Considerations}

Our human evaluation study was conducted with approval from the Institutional Review Board (IRB), ensuring all procedures adhered to established ethical research standards.
All participants recruited for the annotation task were compensated for their time at a rate of 30,000 KRW (approximately 21.76 USD as of September 2025), a rate that meets or exceeds fair compensation guidelines for our region.

\section*{Acknowledgments}
This work was supported by Institute of Information \& communications Technology Planning \& Evaluation (IITP) grant funded by the Korea government (MSIT) (No. RS-2024-00509258 and No. RS-2024-00469482, Global AI Frontier Lab).

This research project has benefitted from the Microsoft Accelerate Foundation Models Research (AFMR) grant program through which leading foundation models hosted by Microsoft Azure along with access to Azure credits were provided to conduct the research.

We acknowledge using ChatGPT\footnote{\url{https://chatgpt.com}} and Claude\footnote{\url{https://claude.ai}} for writing and coding assistance, and Perplexity\footnote{\url{https://perplexity.ai}} and OpenScholar~\citep{asai2024openscholar} for literature search.


\clearpage

\appendix

\section*{Appendix}
\label{sec:appendix}

\section{Data Preparation}
\label{sec:appx_data_prep}

\subsection{Languages}
\label{sec:appx_langs}
Throughout this paper, we evaluated LLMs across 30 languages: 10 high-resource, 10 mid-resource, and 10 low-resource languages. The resource classification follows the categorization defined by~\citet{singh-etal-2024-aya}.

\begin{table}[H]
  \resizebox{\columnwidth}{!}{
    \begin{tabular}{lllc}
      \toprule
      ISO Code                            & Language   & Script           & Resources \\
      \midrule
      \texttt{arb\_Arab} & Arabic     & Arabic         & High      \\
      \texttt{deu\_Latn} & German     & Latin          & High      \\
      \texttt{eng\_Latn} & English    & Latin          & High      \\
      \texttt{fra\_Latn} & French     & Latin          & High      \\
      \texttt{hin\_Deva} & Hindi      & Devanagari     & High      \\
      \texttt{ita\_Latn} & Italian    & Latin          & High      \\
      \texttt{jpn\_Jpan} & Japanese   & Japanese       & High      \\
      \texttt{por\_Latn} & Portuguese & Latin          & High      \\
      \texttt{spa\_Latn} & Spanish    & Latin          & High      \\
      \texttt{zho\_Hans} & Chinese    & Simplified Han & High      \\
      \texttt{ben\_Beng} & Bengali    & Bengali        & Mid       \\
      \texttt{ell\_Grek} & Greek      & Greek          & Mid       \\
      \texttt{heb\_Hebr} & Hebrew     & Hebrew         & Mid       \\
      \texttt{ind\_Latn} & Indonesian & Latin          & Mid       \\
      \texttt{kor\_Hang} & Korean     & Hangul         & Mid       \\
      \texttt{lit\_Latn} & Lithuanian & Latin          & Mid       \\
      \texttt{ron\_Latn} & Romanian   & Latin          & Mid       \\
      \texttt{tha\_Thai} & Thai       & Thai           & Mid       \\
      \texttt{ukr\_Cyrl} & Ukrainian  & Cyrillic       & Mid       \\
      \texttt{zsm\_Latn} & Malay      & Latin          & Mid       \\
      \texttt{amh\_Ethi} & Amharic    & Ethiopic       & Low       \\
      \texttt{hau\_Latn} & Hausa      & Latin          & Low       \\
      \texttt{ibo\_Latn} & Igbo       & Latin          & Low       \\
      \texttt{kir\_Cyrl} & Kyrgyz     & Cyrillic       & Low       \\
      \texttt{npi\_Deva} & Nepali     & Devanagari     & Low       \\
      \texttt{sin\_Sinh} & Sinhala    & Sinhala        & Low       \\
      \texttt{som\_Latn} & Somali     & Latin          & Low       \\
      \texttt{swh\_Latn} & Swahili    & Latin          & Low       \\
      \texttt{tel\_Telu} & Telugu     & Telugu         & Low       \\
      \texttt{yor\_Latn} & Yoruba     & Latin          & Low       \\
      \bottomrule
    \end{tabular}
  }
  \caption{All 30 languages used in this paper with each language's corresponding ISO codes, scripts, and resource classifications defined by~\citet{singh-etal-2024-aya}}
\end{table}

\subsection{Datasets}

\label{sec:appx_datasets}

\paragraph{Easy Twenty Questions.}

We began with 200 English words from the dev and test sets of the \emph{Things}\footnote{\url{https://github.com/apple/ml-entity-deduction-arena}} dataset~\citep{zhang-etal-2024-probing}.
We translated these words into all 30 target languages using Google Translate\footnote{\url{https://translate.google.com}}.
To ensure consistency and quality, we applied several filtering steps: we removed words where Latin characters persisted in non-Latin script languages, eliminated duplicates within each language, and filtered out remaining loan words to ensure semantic consistency across all languages.
This filtering process yielded a final set of 140 words that maintained equivalence across all 30 languages.
For each target word in each language, we randomly sampled 99 additional words from the same language to create a candidate pool of 100 words.
The composition of these candidate pools and their ordering were kept consistent across all languages to ensure fair comparison.
Table~\ref{tab:easy20q-words} provides example target words used in the Easy Twenty Questions task.
\begin{table}[ht!]
  \centering
  \includegraphics[width=\linewidth]{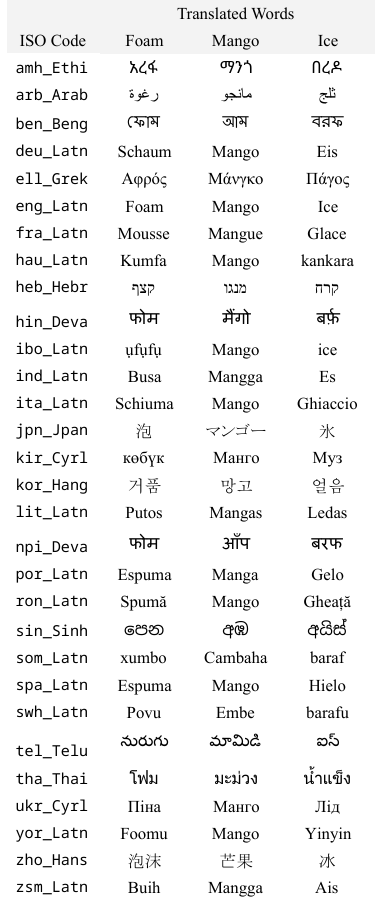}
  \caption{Example target words used in the Easy Twenty Questions task. Words were sourced from the \emph{Things} dataset and translated into 30 languages via Google Translate.}
  \label{tab:easy20q-words}
\end{table}

\paragraph{Other tasks and benchmarks.}
We utilized datasets available on Hugging Face for Belebele\footnote{\label{fn:belebele}\url{https://hf.co/datasets/facebook/belebele}}, HumanEvalExplain\footnote{\url{https://hf.co/datasets/bigcode/humanevalpack}}, Global-MMLU\footnote{\url{https://hf.co/datasets/CohereLabs/Global-MMLU}}, and MultiQ\footnote{\url{https://hf.co/datasets/caro-holt/MultiQ}}.
Our experiments included the same 30 languages for Belebele and MultiQ that we used in our framework, while Global-MMLU experiments covered 29 languages (excluding Thai).
For Global-MMLU, we specifically used only the Culturally-Agnostic (CA) subset to ensure fair cross-lingual comparability across all evaluated languages.

\section{Experimental Setup}
\label{sec:appx_exp_setup}

\subsection{Models}
\label{sec:appx_models}
We conduct our evaluation by selecting recent LLMs, accessing with APIs. This information is summarized in Table~\ref{tab:model_info}.
\begin{table*}[ht]
  \footnotesize
  \centering
  \begin{tabular}{lll}
    \toprule
    Model            & Model Identifier                                            & API Provider                          \\
    \midrule
    GPT-4o           & \texttt{gpt-4o-2024-08-06}                 & \multirow{2}{*}{Azure OpenAI Service} \\
    GPT-4o-mini      & \texttt{gpt-4o-mini-2024-07-18}            &                                       \\
    \midrule
    Gemini-2.5-flash & \texttt{gemini-2.5-flash-preview-04-17}    & \multirow{6}{*}{OpenRouter}           \\
    Gemini-2.0-flash & \texttt{gemini-2.0-flash-001}              &                                       \\
    Qwen2.5-72B      & \texttt{Qwen/Qwen2.5-72B-Instruct}         &                                       \\
    Qwen2.5-7B       & \texttt{Qwen/Qwen2.5-7B-Instruct}          &                                       \\
    Llama-3.3-70B    & \texttt{meta-llama/Llama-3.3-70B-Instruct} &                                       \\
    Llama-3.1-8B     & \texttt{meta-llama/Llama-3.1-8B-Instruct}  &                                      \\
    \bottomrule
  \end{tabular}
  \caption{Model identifiers and API providers used in experiments}
  \label{tab:model_info}
\end{table*}

\subsection{Generations}
\label{sec:appx_generations}

The tasks used in our evaluation were configured with different generation parameters, such as temperature, token limits, and thresholds for fidelity scoring. Details for each task are provided in Table~\ref{tab:task-config}.

\begin{table*}
  \centering
  \footnotesize
  \begin{tabular}{llll}
    \toprule
    Name    & Temperature       & Max Tokens         & Fidelity Threshold    \\
    \midrule
    Easy Twenty Questions & 0.7 &
    \begin{tabular}[c]{@{}l@{}}Questioner: 1024\\ Answerer: 128
    \end{tabular} &
    \begin{tabular}[c]{@{}l@{}}Language: 0.7\\ Answer: 0.9
    \end{tabular} \\
    MCQ conversation & 0.7  &
    \begin{tabular}[c]{@{}l@{}}Questioner: 2048\\ Answerer: 256
    \end{tabular} &
    \begin{tabular}[c]{@{}l@{}}Language: 0.9\\ Answer: 0.9
    \end{tabular} \\
    Code Reconstruction  &
    \begin{tabular}[c]{@{}l@{}}Describer: 0.7\\ Rebuilder: 0.2
    \end{tabular} & 2048  & Language: 0.9 \\
    Global MMLU & 0.0 & 32 & N/A \\
    Belebele & 0.7   & 2048   & N/A  \\
    MultiQ & 0.0                                                                      &
    \begin{tabular}[c]{@{}l@{}}Model: 256\\ Judge: 32
    \end{tabular}           & Language: 0.9
    \\
    \bottomrule
  \end{tabular}
  \caption{Task-specific generation settings used in the evaluation}
  \label{tab:task-config}
\end{table*}

\paragraph{Generation settings.}
We modified several benchmark settings to ensure fair multilingual comparison.
Key adjustments included explicitly prompting models to use the target language, rather than assuming responses would match the question language.
For Code Reconstruction, we removed code description length limits since consistent length constraints across different scripts isn't feasible.
We use 5-shot prompting for Global-MMLU and zero-shot for Belebele.

\paragraph{Prompts.}
We provide prompts used for the three main tasks introduced in Section~\ref{sec:tasks}, as well as for established benchmarks which are Belebele~\citep{bandarkar-etal-2024-belebele}, MultiQ~\citep{holtermann-etal-2024-evaluating}, and Global-MMLU~\citep{singh2024global} (for section \cref{sec:corr}).
Each table outlines the role-specific prompts that we provided to two separate model instances. For Easy Twenty Questions and MCQ Conversation, the instances act as a \textit{questioner} and an \textit{answerer}; for Code Reconstruction, they act as a \textit{describer} and a \textit{rebuilder}.
The prompt for Easy Twenty Questions is provided in Table~\ref{tab:prompt-20q}, MCQ Conversation is in Table~\ref{tab:prompt-mcqc}, and Code Reconstruction is in Table~\ref{tab:prompt-coderebuild}. The prompts for the preexisting three tasks are provided in Table~\ref{tab:prompt-preexisting}.

\paragraph{Cost Analysis.}
The total cost to replicate our main results (Table~\ref{tab:main_table}) was approximately 608 USD, calculated using API pricing from OpenRouter and Azure OpenAI Service.
The costs were distributed across the tasks as follows: Easy Twenty Questions (252 USD), MCQ Conversation (338 USD), and Code Reconstruction (18 USD).
Notably, the evaluation of GPT-4o, our most expensive model, accounted for the majority of this expenditure at 449 USD.

\section{Detailed Experiment Results and Analysis}
\label{sec:appx_more_results}

This section presents a comprehensive breakdown of our experimental results, including task-specific performance and its cross-lingual comparisons across multiple models. We also provide visualizations of task-wise correlations and additional evaluation results not included in the main paper.

\subsection{Results on all languages on all models}
\label{sec:appx_all_result}
Table~\ref{tab:all_result1}, \ref{tab:all_result2} present the evaluation results for all eight models across 30 languages and three tasks.
For each model, we report task-wise accuracy scores across all languages, along with their corresponding Z-scores.

To account for varying task difficulties and enable a unified language ranking per model, we compute Z-scores that aggregate performance across the three tasks.
Each task's scores are standardized independently, using the global mean and standard deviation computed over all models and languages for that task.
This ensures that task-specific differences in difficulty are normalized appropriately.
We then compute the average Z-score across the three tasks per language, allowing for relative performance comparisons across languages within each model.

A Z-score above 0 indicates that the model's accuracy on that language is above the global average, while a negative score suggests below-average performance.
These aggregated Z-scores provide a normalized basis for ranking languages within each model and allow for interpretable comparisons.

\subsection{Visualizations of task-wise correlations}
\label{appx:visual_6by6}

\begin{figure*}[ht]
  \centering
  \includegraphics[width=\linewidth]{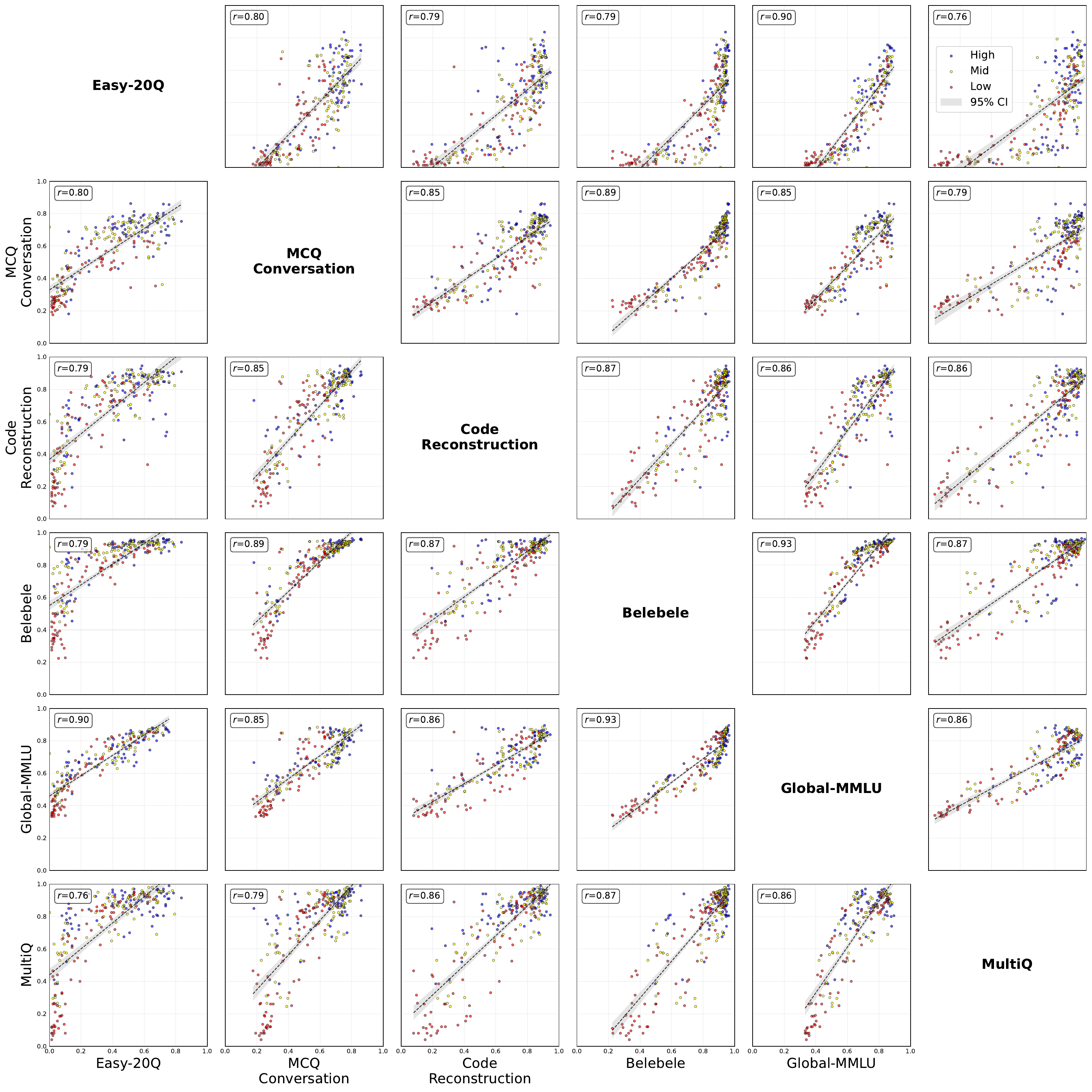}
  \caption{Correlation matrix showing relationships between \tool tasks and existing multilingual benchmarks. Each cell displays Pearson's correlation coefficient ($r$) with 95\% confidence intervals, with points colored by language resource level.}
  \label{fig:corr_scatter}
\end{figure*}

We present a set of $6 \times 6$ scatter plots in Figure~\ref{fig:corr_scatter}, visualizing pairwise correlations between the six tasks. Each plot compares the accuracy scores of two tasks across all 30 languages for 8 models, resulting in one point per language per model.

Each point in a scatter plot represents the performance of a particular language on two different tasks, with the $x$- and $y$-axes indicating the accuracy scores for each task. These visualizations help identify trends and clusters, revealing how performance on one task relates to another across languages.

These scatter plots serve as a visual counterpart to the Pearson correlation coefficients ($r$) reported in Figure~\ref{fig:comparision_benchmarks}, offering an intuitive understanding of inter-task relationships observed in our experiments.

\subsection{Additional plot about language resource flexibility on MCQ Conversation}
\label{sec:appx_mini_plot_result}
Following up on the analysis in Section~\ref{sec:dollar}, we conducted the same experiment with GPT-4o-mini under identical settings.

\begin{figure}[H]
  \centering
  \includegraphics[width=0.9\linewidth]{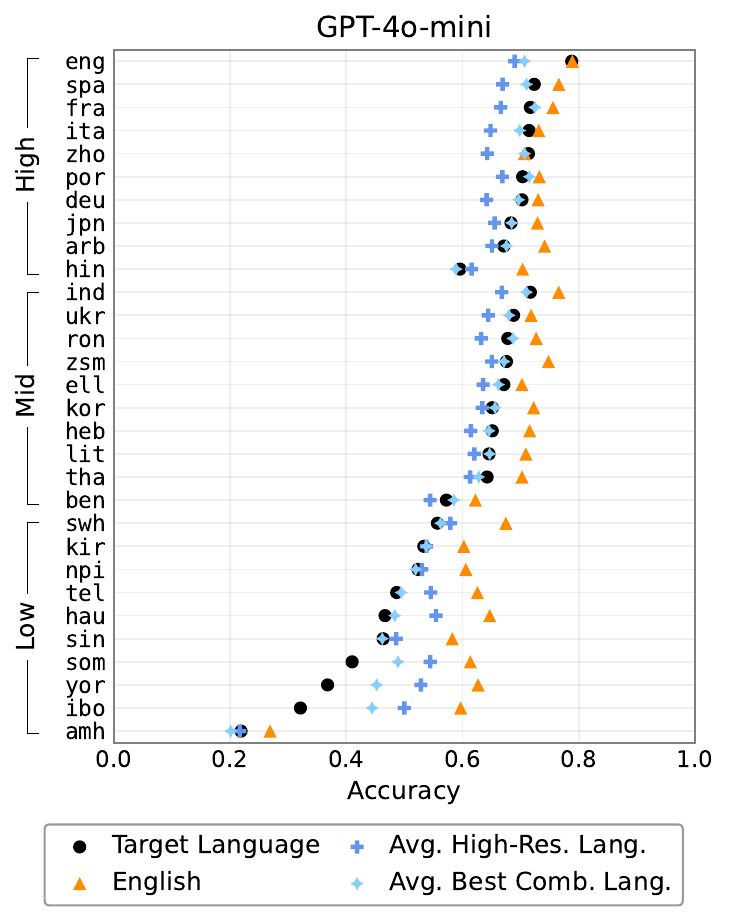}
  \caption{MCQ Conversation accuracy comparison across 30 languages for GPT-4o-mini, using passages in:
    (1) the target language,
    (2) English,
    (3) a fixed set of five high-resource languages (averaged), and
  (4) a selection of up to five high-resource languages most similar to the target language, with scores averaged.}
  \label{fig:dollar_index_mini}
\end{figure}

Figure~\ref{fig:dollar_index_mini} presents the MCQ Conversation accuracy across 30 languages when passages are provided in four different conditions: (1) the target language, (2) English, (3) a fixed set of five high-resource languages (averaged), and (4) a selection of up to five high-resource languages that are most similar to the target language. The overall trend is consistent with that of GPT-4o (Figure~\ref{fig:dollar_index}).

\subsection{Additional analysis about language resource flexibility on MCQ Conversation}
\label{appx:appx_mcqconv_comb}

To complement the substitution analysis in Section~\ref{sec:dollar}, Table~\ref{tab:dollar_lang} lists, for each of the 30 target languages, the subset of high-resource languages (selected from English, Chinese, Japanese, Hindi, and Arabic) that most closely approximates the original target-language passage in terms of MCQ Conversation accuracy.

The optimal subset for each target language was determined by selecting the combination (up to five languages) that minimizes the L2 distance from the original accuracy, as described in Section~\ref{sec:dollar}. When the target language itself was one of the five high-resource languages, it was excluded from its own substitution set. These exclusions are marked with \xmark\ in the corresponding table entries.
\begin{table}[H]
  \resizebox{\columnwidth}{!}{%
    \begin{tabular}{@{}llcccccc@{}}
      \toprule
      ISO Code  & Language        & Resources & \textsc{Eng} & \textsc{Zho} & \textsc{Arb} & \textsc{Jpn} & \textsc{Hin} \\ \midrule
      \texttt{spa\_Latn} & Spanish & High & \ding{52} & \ding{52} & \ding{52} &  &  \\
      \texttt{arb\_Arab} & Arabic & High & \ding{52} & \ding{52} & \xmark & \ding{52} &  \\
      \texttt{deu\_Latn} & German & High & \ding{52} & \ding{52} &  &  &  \\
      \texttt{fra\_Latn} & French & High & \ding{52} & \ding{52} &  &  &  \\
      \texttt{ita\_Latn} & Italian & High & \ding{52} & \ding{52} &  &  &  \\
      \texttt{por\_Latn} & Portuguese & High & \ding{52} & \ding{52} &  &  &  \\
      \texttt{zho\_Hans} & Chinese & High & \ding{52} & \xmark &  &  &  \\
      \texttt{eng\_Latn} & English & High & \xmark & \ding{52} &  &  &  \\
      \texttt{jpn\_Jpan} & Japanese & High &  & \ding{52} &  & \xmark &  \\
      \texttt{hin\_Deva} & Hindi & High &  &  &  & \ding{52} & \xmark \\ \midrule
      \texttt{zsm\_Latn} & Malay & Mid & \ding{52} & \ding{52} & \ding{52} & \ding{52} &  \\
      \texttt{lit\_Latn} & Lithuanian & Mid & \ding{52} & \ding{52} & \ding{52} &  &  \\
      \texttt{kor\_Hang} & Korean & Mid & \ding{52} &  & \ding{52} & \ding{52} &  \\
      \texttt{ben\_Beng} & Bengali & Mid & \ding{52} & \ding{52} &  &  &  \\
      \texttt{ron\_Latn} & Romanian & Mid & \ding{52} & \ding{52} &  &  &  \\
      \texttt{ukr\_Cyrl} & Ukrainian & Mid & \ding{52} &  & \ding{52} &  &  \\
      \texttt{ell\_Grek} & Greek & Mid & \ding{52} &  &  & \ding{52} &  \\
      \texttt{heb\_Hebr} & Hebrew & Mid & \ding{52} &  &  & \ding{52} &  \\
      \texttt{ind\_Latn} & Indonesian & Mid & \ding{52} &  &  & \ding{52} &  \\
      \texttt{tha\_Thai} & Thai & Mid & \ding{52} &  &  &  & \ding{52} \\ \midrule
      \texttt{sin\_Sinh} & Sinhala & Low &  & \ding{52} & \ding{52} & \ding{52} & \ding{52} \\
      \texttt{npi\_Deva} & Nepali & Low & \ding{52} &  & \ding{52} &  & \ding{52} \\
      \texttt{kir\_Cyrl} & Kyrgyz & Low &  & \ding{52} & \ding{52} &  &  \\
      \texttt{amh\_Ethi} & Amharic & Low &  &  & \ding{52} &  &  \\
      \texttt{swh\_Latn} & Swahili & Low &  &  & \ding{52} &  &  \\
      \texttt{hau\_Latn} & Hausa & Low &  &  &  &  & \ding{52} \\
      \texttt{ibo\_Latn} & Igbo & Low &  &  &  &  & \ding{52} \\
      \texttt{som\_Latn} & Somali & Low &  &  &  &  & \ding{52} \\
      \texttt{tel\_Telu} & Telugu & Low &  &  &  &  & \ding{52} \\
      \texttt{yor\_Latn} & Yoruba & Low &  &  &  &  & \ding{52} \\ \bottomrule
    \end{tabular}%
  }
  \caption{Optimal subsets of high-resource languages (selected from English, Chinese, Japanese, Hindi, and Arabic) for approximating the native-language passage performance in the MCQ Conversation task.
    For each target language, the listed subset scores the lowest L2 distance from the original accuracy.
  If the target language is one of the five high-resource options, it is excluded from its own substitution set, denoted with \xmark\ }
  \label{tab:dollar_lang}
\end{table}

\subsection{Human analysis case on MCQ Conversation Errors}
\label{appx:quali}
As described in Section~\ref{sec:quali}, we conducted a qualitative error analysis for both the Easy Twenty Questions and MCQ Conversation tasks.
Specifically, we examined which conversational agent—the Questioner or the Answerer—was primarily responsible for task failure in each case.
Tables~\ref{tab:q_wrong_case} and~\ref{tab:a_wrong_case} provide illustrative examples of typical errors for each role, along with our analysis of the underlying issues.

\subsection{Correlation with Human Evaluation on MultiQ}
To further validate \tool{}'s effectiveness as a proxy for human evaluation, we conducted a human analysis for MultiQ dataset on 14 languages and 8 models.

\subsubsection{Setup}
To empirically validate \tool's automated scores against human judgments, we conduct a human evaluation study.
Because the conversational logs from our framework are highly structured, we use the more open-ended MultiQ benchmark~\cite{holtermann-etal-2024-evaluating} to test whether \tool scores generalize as a reliable proxy for general-purpose text quality.
This study evaluates outputs from the eight LLMs for a set of 15 questions sampled from the original 200 in the MultiQ benchmark.
The evaluation spans 14 languages selected to cover high-resource (Arabic, Chinese, English, French, Hindi), mid-resource (Bengali, Indonesian, Korean, Malay, Thai), and low-resource (Amharic, Kyrgyz, Sinhala, Swahili) categories.

We recruit 12 annotators---primarily university students in South Korea---each a native speaker of their assigned language(s).
Ten annotators cover a single language, while two bilingual annotators are responsible for two languages each (French/Arabic and Chinese/Malay).
The same 15 questions are selected for all languages.
To ensure the evaluation is both manageable and effective at differentiating model performance, we prioritize the most challenging questions based on a preliminary LLM-as-judge scoring using Gemini-2.5-flash.
Before the main task, annotators are calibrated using a standardized set of English examples scored by the authors to ensure consistent judgment.
Each participant evaluates the full set of generated responses for their language(s) in a two-hour session.
Following a rubric adapted from \citet{hada-etal-2024-metal}, responses are scored on a 5-point Likert scale across three criteria: Linguistic Acceptability (fluency and naturalness), Output Content Quality (coherence and clarity), and Task Quality (how well the response addresses the question).

\subsubsection{Result}
\begin{figure}[ht]
  \centering
  \includegraphics[width=\linewidth]{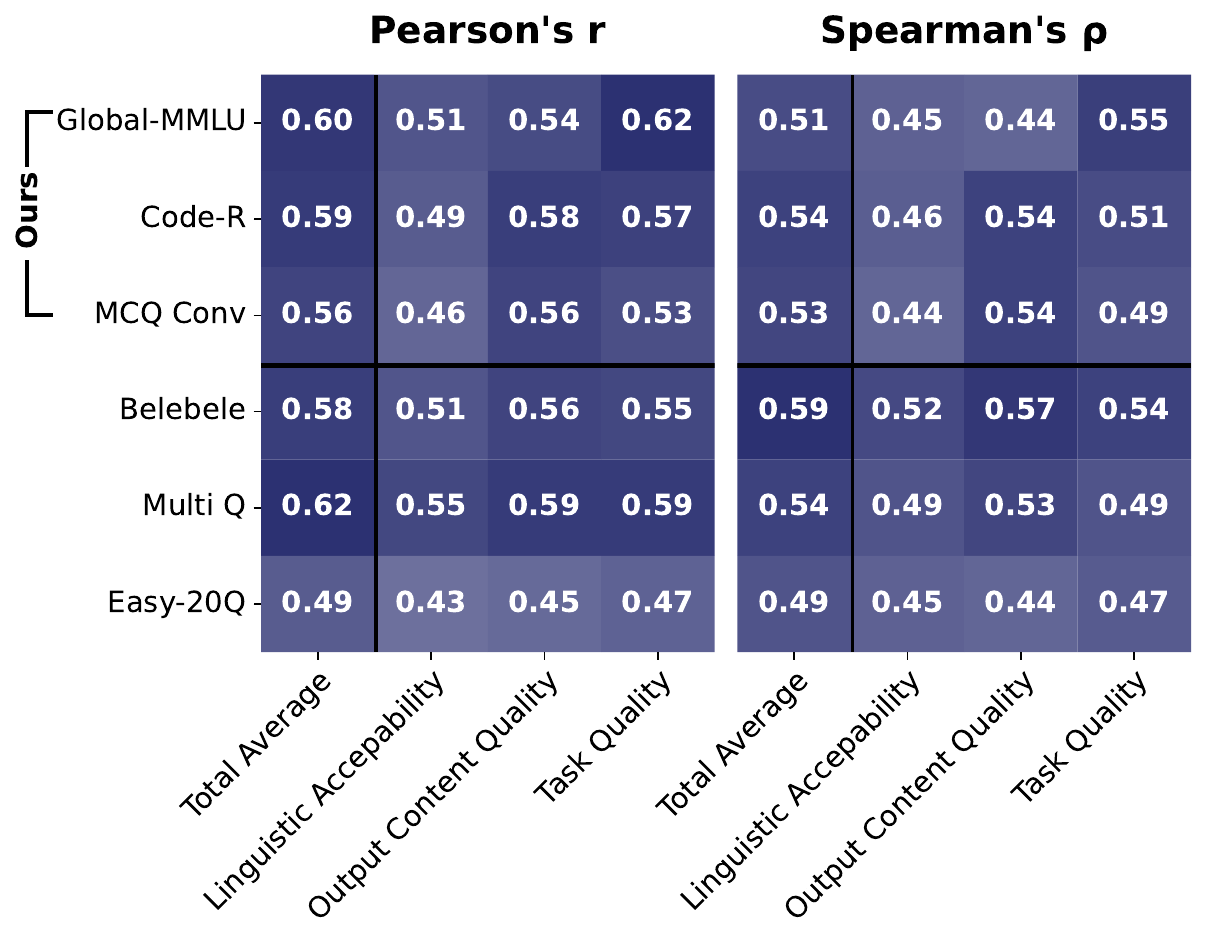}
  \caption{Correlation analysis between human annotation on MultiQ data and six tasks consisting of \tool{} and existing multilingual benchmarks. Heatmaps show Pearson's $r$ (left) and Spearman's $\rho$ (right) correlation coefficients between human annotation and six tasks. All correlations exceed 0.4, demonstrating medium to strong consistency between human annotation with other six tasks, validating \tool{}'s effectiveness as a multilingual evaluation framework.}
  \label{fig:human_correlation_heatmap}
\end{figure}

For each question–answer set across 14 languages and 8 models, we first computed annotation scores for three metrics—Linguistic Acceptability, Output Content Quality, and Task Quality—and then averaged them to obtain a Total Average score per sample. These final annotation scores were subsequently averaged across samples for each language–model pair, yielding 112 aggregated scores (14 languages × 8 models).
We then examined the correlation between these human evaluation scores (based on the three criteria) and task-specific scores from \tool{} as well as three existing multilingual benchmarks, all provided per language and model. Correlation statistics are reported in Figure~\ref{fig:human_correlation_heatmap}.

The results show moderate to strong correlations between human judgments and \tool{} scores across tasks and metrics. This demonstrates that although \tool{} was originally designed for structured, information-gap tasks, its task completion–based scores generalize well to open-ended question answering in MultiQ. The strongest correlation was with Task Quality and the weakest with Linguistic Acceptability, reflecting \tool{}'s focus on accurate information transfer rather than fluency. These findings suggest that \tool{} scores align well with human evaluation, though the three metrics are not fully independent.

\subsection{Extending \tool{} to Summarization}
To demonstrate the extensibility of the \tool framework beyond our initial three tasks, we implement a new summarization task based on the same ``information-gap'' paradigm underlying \tool's design.

\subsubsection{Methodology}
We employed the same 8 models and 30 languages from \tool. The summarization evaluation was conducted as follows:

\textbf{Summarization-Length Limit Normalization:} Using the FLORES+~\citep{goyal-etal-2022-flores} dataset, which primarily contains human-translated texts, we sampled 100 English sentences and retrieved their translations in 30 target languages. For each pair, we computed the ratio of character lengths by dividing the length of the translated sentence in a target language by the length of the corresponding English sentence, using Python's \texttt{len()} function. These ratios were then applied for length control in multilingual summarization.

\textbf{Dataset:} We sampled 100 articles from the QAGS~\citep{wang-etal-2020-asking} dataset, each originally in English. For each article, we generated 5 English question–answer pairs using GPT-4o-mini, with answers reflecting key factual entities.

\textbf{Evaluation Process:} For each model, language, and article, we followed this process:
\begin{enumerate}
  \item A Summarizer LLM (target model) produced a summary of the article in the target language. Its language was verified using GlotLID, and the length was constrained to \((\texttt{Original English article length}) \times 0.5 \times (\texttt{language length ratio})\).
  \item An Answerer LLM (target model) received only the target-language summary and the English questions, and generated answers in English.
  \item The generated answers were compared to the gold answers using an LLM-as-Judge (GPT-4o-mini). Since both gold and generated answers were in English, this evaluation setup avoids translation-related bias.
\end{enumerate}

\subsubsection{Results}
The average accuracy of the Answerer LLM across all models and languages is reported in Table~\ref{tab:summary_result}, and its correlation score with six tasks consisting of \tool{} and existing multilingual benchmarks is reported in Table~\ref{tab:summary_correlation}. This experiment shows that \tool{} can be readily extended to summarization while preserving its information-gap design, enabling scalable evaluation without references or human judgments. The results further exhibit moderate-to-strong correlations with the original \tool{} tasks, indicating that the framework captures a generalizable signal of multilingual generation quality.

\section{Generation Statistics}
\label{appdx:stats}
As stated in Section~\ref{sec:stats}, we report detailed generation statistics in Table~\ref{tab:stats}, averaged over models and language groups. Specifically, we measured the following:

\begin{itemize}
  \item \textbf{Token Count and Sequence Length}: The number of tokens (\# Token) and total character count (\# Char) are computed from outputs generated in the target language by the questioner or the describer. The number of tokens were computed using the tokenizer associated with each model used in the experiments.
  \item \textbf{Language Fidelity}: Fidelity is measured as the percentage of questioner or describer outputs identified by GlotLID as matching the target language.
  \item \textbf{Instruction-Following of the Answerer}: Answerer Instruction-Following (A I-F) is defined for Easy Twenty Questions and MCQ Conversation as the proportion of answerer responses that strictly follow the output format (``yes,'' ``no,'' and ``maybe'').
  \item \textbf{Interaction Length}: The number of question turns per interaction (\# Turn) is reported for Easy Twenty Questions and MCQ Conversation, both of which are multi-turn tasks.
\end{itemize}

\begin{table}[H]
  \resizebox{\columnwidth}{!}{
    \begin{tabular}{lcc}
      \toprule
      & Pearson & Spearman \\
      \midrule
      Easy Twenty Questions & 0.65    & 0.63     \\
      MCQ Conversation      & 0.65    & 0.61     \\
      Code Reconstruction   & \textbf{0.79}    & \textbf{0.74}     \\
      Global MMLU           & 0.68    & 0.68     \\
      Belebele              & 0.66    & 0.65     \\
      MultiQ                & 0.62    & 0.65     \\
      \bottomrule
    \end{tabular}
  }
  \caption{Correlation score of summarization task with six tasks consisting of \tool{} and existing multilingual benchmarks. Overall correlation scores show high correlation, suggesting that the extension of \tool{} to other domains is plausible.}
  \label{tab:summary_correlation}
\end{table}

\begin{table*}[ht]
  \centering
  \resizebox{\textwidth}{!}{%
\begin{tabular}{@{}lr|ccccc|ccccc|ccc@{}}
      \toprule
      \multicolumn{2}{c|}{\multirow{2}{*}{}}& \multicolumn{5}{c|}{Easy Twenty Questions}      & \multicolumn{5}{c|}{MCQ Conversation}  & \multicolumn{3}{c}{Code Reconstruction} \\
      \multicolumn{2}{c|}{}       & \# Token & \# Char. & Fidelity & A I-F   & \# Turn & \# Token & \# Char. & Fidelity & A I-F   & \# Turn & \# Token    & \# Char.    & Fidelity    \\
      \midrule
      \multicolumn{1}{c}{\multirow{5}{*}{\rotatebox{90}{Language}}} & All              & 29.95    & 52.12    & 95.00    & 99.57  & 14.33 & 49.07    & 103.85   & 98.32    & 99.50  & 3.99  & 181.04      & 374.30      & 97.72       \\
      \multicolumn{1}{c}{}      & \textsc{Eng}              & 11.19    & 45.28    & 96.52    & 100.00 & 14.32 & 23.22    & 111.43   & 99.88    & 99.95  & 4.05  & 93.50       & 412.86      & 99.63       \\
      \multicolumn{1}{c}{}      & High             & 16.19    & 44.47    & 95.78    & 99.30  & 14.25 & 30.24    & 94.63    & 97.76    & 99.37  & 3.98  & 113.36      & 341.29      & 97.91       \\
      \multicolumn{1}{c}{}      & Mid              & 18.88    & 41.97    & 95.59    & 99.53  & 14.32 & 38.16    & 92.54    & 98.93    & 99.72  & 3.95  & 147.26      & 344.32      & 98.64       \\
      \multicolumn{1}{c}{}      & Low              & 54.77    & 69.87    & 93.61    & 99.88  & 14.42 & 78.70    & 124.31   & 98.28    & 99.40  & 4.04  & 282.50      & 437.31      & 96.61       \\
      \midrule
      \multirow{8}{*}{\rotatebox{90}{Model}}    & GPT-4o           & 14.80    & 38.52    & 97.16    & 100.00 & 13.96 & 27.10    & 71.54    & 99.80    & 100.00 & 4.02  & 123.68      & 345.46      & 99.91       \\
\bottomrule
    \end{tabular}%
  }
  \caption{Average token count (\# Token), character-level sequence length (\# Character), GlotLID-based language fidelity (Fidelity), instruction-following rate of the answerer (A I-F), and average number of question turns (\# Turn) are computed per task, model, and language group.}
  \label{tab:stats}
\end{table*}

\newcolumntype{Y}{>{\centering\arraybackslash}p{2cm}}
\begin{table*}[ht]
  \centering
  \resizebox{\textwidth}{!}{
    \begin{tabular}{r|YYYYYYYY}
      \toprule
      \multicolumn{1}{l|}{} & gpt-4o        & gpt-4o-mini   & gemini-2.5-flash & gemini-2.0-flash & qwen-2.5-72b  & qwen-2.5-7b   & llama-3.3-70b & llama-3.1-8b  \\
      \midrule
      Spanish               & 0.76          & 0.77          & 0.78             & \textbf{0.79}    & 0.81          & 0.62          & \textbf{0.52} & 0.4           \\
      Arabic                & 0.57          & 0.56          & 0.54             & {\ul 0.53}       & 0.57          & 0.3           & {\ul 0.25}    & 0.07          \\
      German                & 0.76          & \textbf{0.78} & 0.8              & 0.75             & 0.81          & 0.55          & 0.49          & 0.39          \\
      French                & 0.76          & 0.76          & 0.79             & 0.76             & 0.8           & 0.6           & 0.46          & \textbf{0.46} \\
      Italian               & 0.76          & 0.75          & 0.8              & 0.78             & 0.8           & 0.6           & 0.45          & 0.34          \\
      Portuguese            & \textbf{0.79} & 0.77          & 0.78             & 0.74             & 0.82          & 0.57          & 0.5           & 0.39          \\
      Chinese               & 0.56          & 0.56          & 0.6              & 0.63             & 0.75          & 0.56          & 0.36          & 0.18          \\
\bottomrule
    \end{tabular}
  }
  \caption{Results of the summarization task: average accuracy of the Answerer LLM across 8 models and 30 languages, demonstrating the extensibility of \tool{} framework}
  \label{tab:summary_result}
\end{table*}

\begin{table*}[ht]
  \resizebox{\textwidth}{!}{
    \begin{tabular}{p{2cm} p{2cm} p{12cm}}
      \toprule
      Role & Type & Prompt \\
      \midrule
      Questioner & Initial \newline Instruction &
      You will be solving an entity deduction game by asking questions about a hidden item in \{lang\_full\}. Your goal is to identify exactly one correct entity from a list of 100 items through strategic questioning, using as few questions as possible. You can ask yes/no/maybe questions in \{lang\_full\}, one at a time. Each question must be concise and answerable only with "Yes," "No," or "Maybe." Do not ask for hints or the answer directly. \newline

      Among the list, exactly one is correct. You have a maximum of 20 questions before making your final choice, but fewer questions are better. When you have determined the correct entity, provide your final answer using double brackets: {[}{[}entity{]}{]}. The entity must be exactly as it appears in the Entity List (do not translate to another language).\newline

      \textless{}|Entity List|\textgreater \newline
      \{entity\_list\} \newline

      \textless{}|Language|\textgreater \newline
      \{lang\_full\} \newline

      Begin with your first question. \\
      \midrule

      Answerer & Instruction  &  "Answer the following question about the hidden entity: \( \{entity\} \). Respond only with ""Yes,"" ""No,"" or ""Maybe"" (including the period) without providing explanations or additional words. Do not use ""Maybe"" if the answer can be definitively ""Yes"" or ""No."" Your responses must be in English. Never mention the entity in your response. If the query contains two or more questions asked simultaneously, consider it invalid and respond with ""Maybe.""\newline

      <|Entity|>\newline
      \( \{entity\} \)\newline

      <|Language|>\newline
      \( \{lang\_full\} \)\newline

      <|Question|>\newline
      \( \{question\} \)\newline

      <|Options|>\newline

      - Yes.\newline
      - No.\newline
      - Maybe.
      \\
\bottomrule
    \end{tabular}
  }
  \caption{Prompt design for the Easy Twenty Questions task. The questioner and answerer are separately prompted with role-specific instructions to simulate a Twenty Questions game. The prompts include task rules, language constraints, response formatting requirements, and structured input fields (\eg \texttt{\{entity\_list\}}, \texttt{\{lang\_full\}}, \texttt{\{question\}})}
  \label{tab:prompt-20q}
\end{table*}

\begin{table*}[ht]
  \centering
  \resizebox{0.9\textwidth}{!}{
    \begin{tabular}{p{2cm} p{2cm} p{12cm}}
      \toprule
      Role       & Type                                                                         & Prompt \\
      \midrule
      Questioner & Initial \newline Instruction & You will be solving a multiple-choice question by asking questions about a hidden passage. I am serving as an intermediary between you and a person who has this passage. You can ask me questions about the passage content, and I will relay these to the person. They will respond with only ``yes,'' ``no,'' or ``maybe,'' which I will then share with you.\newline

      Your questions must be in \{lang\_full\} and you can only ask one question at a time. Do not ask for hints or request the passage directly. Among the four answer choices provided, exactly one is correct. You must ask exactly 4 questions (one corresponding to each answer choice) before making your final decision. After receiving all four responses, provide your final answer in \{lang\_full\}, indicating the correct number choice with double brackets: [[X]]\newline

      <|Query|>\newline
      \{query\}\newline

      <|Choices|>\newline
      (1) \{c1\}\newline
      (2) \{c2\}\newline
      (3) \{c3\}\newline
      (4) \{c4\}\newline

      <|Language|>\newline
      \{lang\_full\} \\
\bottomrule
    \end{tabular}
  }
  \caption{Prompt design for the MCQ Conversation task. The questioner and answerer are assigned separate prompts to simulate a collaborative multiple-choice reasoning task. The questioner asks yes/no questions based on a hidden passage, while the answerer responds with constrained answers. Prompts include language and formatting instructions, as well as structured fields such as \texttt{\{query\}}, \texttt{\{passage\}}, and \texttt{\{lang\_full\}}.}
  \label{tab:prompt-mcqc}
\end{table*}

\begin{table*}[ht]
  \resizebox{\textwidth}{!}{
    \begin{tabular}{p{2cm} p{2cm} p{12cm}}
      \toprule
      Role       & Type                                                                              & Prompt                                                              \\
      \midrule
      Describer & Instruction for \newline the English \newline Language  & Provide a concise natural language description of the code in \( \{ \text{lang\_full} \} \).\newline

      <|Code|>\newline
      \( \{ \text{code} \} \)\newline

      <|Language|>\newline
      \( \{ \text{lang\_full} \} \) \\
      \midrule
      Describer & Instruction for \newline Non-English \newline Languages & Provide a concise natural language description of the code in {lang\_full}. Variable names or technical terms in English are allowed, but there shouldn't be any phrases or full sentences written in English or any language except {lang\_full}.\newline

      <|Code|>\newline
      {code}\newline

      <|Language|>\newline
      {lang\_full}\\
      \midrule
      Rebuilder  & Instruction   & "Write functional code in \( \{code\_lang\_full\} \) according to the description written in \( \{lang\_full\} \). Start your code with the given declaration. Write only the code without any additional descriptions, comments, or delimiters.\newline

      <|Language|>\newline
      \( \{lang\_full\} \)\newline

      <|Description|>\newline
      \( \{description\} \)\newline

      <|Declaration|>\newline
      \( \{declaration\} \) \\
      \bottomrule
    \end{tabular}
  }
  \caption{Prompt design for the Code Reconstruction task. One model instance (the describer) generates a natural language description of a given code snippet in the target language. Another instance (the rebuilder) reconstructs the original function from this description and a given declaration. The prompts specify language requirements and restrict the output format to code-only.}
  \label{tab:prompt-coderebuild}
\end{table*}

\begin{table*}[ht]
  \centering
  \resizebox{\textwidth}{!}{
    \begin{tabular}{p{2cm} p{2cm} p{12cm}}
      \toprule
      Benchmark   & Type & Prompt  \\
      \midrule
      MultiQ & Evaluatee & Please provide an answer to the following question in \( \{ \text{lang\_full} \} \). \newline

      <|Question|> \newline
      \( \{ \text{question} \} \) \newline

      <|Language|> \newline
      \( \{ \text{lang\_full} \} \) \\
      \midrule
      MultiQ      & Classifier \newline (LLM-as-a-Judge) & Determine if the answer to the following question is correct. The question is in English and the answer is in \{lang\_full\}. Respond only with 'yes' or 'no' - do not include explanations or additional words.\newline

      <|Question|>\newline
      \{question\_en\}\newline

      <|Language|>\newline
      \{lang\_full\}\newline

      <|Answer|>\newline
      \{model\_pred\} \\
      \midrule
      Global-MMLU & -  & \{question\}\newline
      A. \{option\_a\}\newline
      B. \{option\_b\}\newline
      C. \{option\_c\}\newline
      D. \{option\_d\}\newline
      Answer:  \\
      \midrule
      Belebele & - & Given the following passage, query, and answer choices, output the number corresponding to the correct answer in double brackets: [[X]]\newline

      <|Language|>\newline
      \{lang\_full\}\newline

      <|Passage|>\newline
      \{passage\}\newline

      <|Query|>\newline
      {query}\newline

      <|Choices|>\newline
      (1) \{c1\}\newline
      (2) \{c2\}\newline
      (3) \{c3\}\newline
      (4) \{c4\} \\
      \bottomrule
    \end{tabular}
  }
  \caption{Prompt design for the pre-existing benchmark tasks used in our evaluation. For MultiQ, we include both evaluatee prompts and classification prompts for LLM-as-a-Judge. Global-MMLU and Belebele use simpler one-shot prompts formatted according to their original task definitions. Prompts include structured input fields such as \texttt{\{question\}}, \texttt{\{lang\_full\}}, and \texttt{\{choices\}}}
  \label{tab:prompt-preexisting}
\end{table*}

\begin{table*}[ht]
  \centering
  \resizebox{\textwidth}{!}{
    \begin{tabular}{p{2cm} p{14cm}}
      \toprule
      Language & Korean \\
      \midrule
      Passage &
      Victoria Falls is a small city in western Zimbabwe, across the border from Livingston, Zambia and Botswana. This town is located right next to the waterfalls, and they are the town's main attraction, and this popular tourist attraction also offers many opportunities for adventurers and tourists alike to stay longer. During the rainy season (November to March), waterfalls increase in volume, and the waterfalls become more dramatic. Crossing a bridge or walking down a winding path near the waterfall will cause your clothes to get wet. On the other hand, the amount of water is so large that the actual size of the waterfall is obscured by the sheer volume! \newline
      {\color{gray}빅토리아 폴스는 리빙스턴, 잠비아와 보츠와나 근처의 국경 건너편의 짐바브웨 서쪽에 있는 작은 도시입니다. 이 마을은 폭포 바로 옆에 위치해 있고, 그 폭포들이 마을의 주요 명소이며, 또한 이 유명한 관광지는 모험가들과 관광객들 모두에게 더 오래 머물 수 있는 많은 기회를 제공한다. 우기(11월부터 3월까지)에는 수량이 증가하며, 폭포가 더 극적으로 변합니다. 다리를 건너거나 폭포 근처의 구불구불한 길을 따라 걸어가면 옷이 젖게 됩니다. 반면에, 물의 양이 너무 많아서 실제 보이는 폭포의 크기는 - 엄청난 수량으로 인해 - 가려진다!} \\
      \midrule
      Query & If I want an \colorbox{red!10}{open view}, when should I visit the waterfall? \newline
      {\color{gray}탁 트인 전망을 원한다면 언제 폭포를 방문해야 합니까?} \\
      \midrule
      Choices &
      1. From November to March {\color{gray}11월부터 3월까지} \newline
      2. During the tourist season {\color{gray}관광철에}\newline
      3. When the water volume is high {\color{gray}수량이 많을 때}\newline
      4. When it's not rainy season {\color{gray}장마철이 아닐 때}
      \\
\bottomrule
    \end{tabular}
  }
  \caption{Human analysis of a case where the Questioner made an erroneous output in MCQ Conversation, ran by \texttt{gpt-4o-2024-08-06}. The original dataset and conversation are in Korean, with the Korean text shown in {\color{gray}gray}.}
  \label{tab:q_wrong_case}
\end{table*}

\begin{table*}[ht]
  \centering
  \resizebox{\textwidth}{!}{
    \begin{tabular}{p{2cm} p{14cm}}
      \toprule
      Language & English \\
      \midrule
      Passage &
      During the 18th century Cambodia found itself squeezed between two powerful neighbors, Thailand and Vietnam. The Thais invaded Cambodia several times in the 18th century and in 1772 they destroyed Phnom Phen. In the last years of the 18th century the Vietnamese also invaded Cambodia. \\
      \midrule
      Query & Which of the following statements accurately describes the invasions of Cambodia? \\
      \midrule
      Choices &
      1. The Vietnamese invaded Cambodia several times \newline
      2. The Thais and Vietnamese invaded Cambodia in 1772 \newline
      3. The Vietnamese destroyed Phnom Phen \newline
      4. The Thais invaded Cambodia several times
      \\
      \midrule
      Conversation &
      Questioner: Is it true that the Vietnamese invaded Cambodia several times?\newline
      Answerer: No.\newline

      Questioner: Did the Thais and Vietnamese invade Cambodia in 1772?\newline
      Answerer: \colorbox{red!10}{Yes.}\newline

      Questioner: Is it true that the Vietnamese destroyed Phnom Penh?\newline
      Answerer: No.\newline

      Questioner: Did the Thais invade Cambodia several times?\newline
      Answerer: Yes. \\
      \midrule
      Error \newline Analysis & The answerer gave an incorrect response to the second question. While there is a mention that the Thais invaded in 1772, there is \textit{no information indicating that the Vietnamese invaded Cambodia in 1772}. However, the answerer responded \colorbox{red!10}{Yes.} to the second question.
      \\
      \bottomrule
    \end{tabular}
  }
  \caption{Human analysis of a case where the Answerer made an erroneous output in MCQ Conversation, ran by \texttt{gpt-4o-2024-08-06}. }
  \label{tab:a_wrong_case}
\end{table*}

\clearpage
\onecolumn

\begin{table}[t]
  \centering
  \resizebox{0.9\textwidth}{!}{
    \begin{tabular}{@{}ccclll|ccclll@{}}
      \toprule
      \multicolumn{6}{c|}{\large \textbf{gpt-4o-2024-08-06}}                                                                                                                  & \multicolumn{6}{c}{\large \textbf{gpt-4o-mini-2024-07-18}}                                                                                                                                                    \\
      &                        & Total    & \multicolumn{1}{c}{E-20Q}      & \multicolumn{1}{c}{MCQ-C}    & \multicolumn{1}{c|}{CR}   & \multicolumn{1}{l}{}                           & \multicolumn{1}{l}{}                       & Total    & \multicolumn{1}{c}{E-20Q}      & \multicolumn{1}{c}{MCQ-C}    & \multicolumn{1}{c}{CR}   \\
      \multirow{-2}{*}{resource}  & \multirow{-2}{*}{lang} & Z avg.  & \multicolumn{1}{c}{Z (Acc)} & \multicolumn{1}{c}{Z (Acc)} & \multicolumn{1}{c|}{Z (Acc)} & \multicolumn{1}{l}{\multirow{-2}{*}{resource}} & \multicolumn{1}{l}{\multirow{-2}{*}{lang}} & Z avg.  & \multicolumn{1}{c}{Z (Acc)} & \multicolumn{1}{c}{Z (Acc)} & \multicolumn{1}{c}{Z (Acc)} \\ \hline
      {\color[HTML]{3531FF} high} & \texttt{eng}                    & 1.37     & 1.62 (75.7) & 1.56 (85.6) & 0.94 (88.4)                    & {\color[HTML]{3531FF} high}                    & \texttt{eng}                                        & 0.94     & 0.7 (53.6) & 1.2 (78.8) & 0.92 (87.8)                    \\
      {\color[HTML]{3531FF} high} & \texttt{zho}                    & 1.33     & 1.94 (83.6) & 1.01 (75.2) & 1.04 (90.9)                    & {\color[HTML]{3531FF} high}                    & \texttt{zho}                                        & 0.77     & 0.76 (55) & 0.8 (71.3) & 0.74 (83.5)                     \\
\bottomrule
    \end{tabular}
  }

  \caption{Results for each task on \tool across 30 languages, evaluated using gpt-4o-2024-08-06, gpt-4o-mini-2024-07-18, gemini-2.5-flash-preview, and gemini-2.0-flash-001. Accuracy was normalized using Z-scores and averaged across tasks. Languages were then ranked by their averaged Z-score.}
  \label{tab:all_result1}
\end{table}

\twocolumn

\clearpage

\begin{table*}[t]
  \centering
  \resizebox{0.9\textwidth}{!}{
    \begin{tabular}{@{}ccclll|ccclll@{}}
      \toprule
      \multicolumn{6}{c|}{\large \textbf{llama-3.3-70b-instruct}}                                                                                                           & \multicolumn{6}{c}{ \large \textbf{llama-3.1-8b-instruct}}                                                                                                                                                    \\
      \multirow{2}{*}{resource} & \multirow{2}{*}{lang} & Total  & \multicolumn{1}{c}{E-20Q}    & \multicolumn{1}{c}{MCQ-C}  & \multicolumn{1}{c|}{CR} & \multicolumn{1}{l}{\multirow{2}{*}{resource}} & \multicolumn{1}{l}{\multirow{2}{*}{lang}} & Total  & \multicolumn{1}{c}{E-20Q}    & \multicolumn{1}{c}{MCQ-C}  & \multicolumn{1}{c}{CR} \\
      &                       & Z avg. & \multicolumn{1}{c}{Z (Acc)} & \multicolumn{1}{c}{Z (Acc)} & \multicolumn{1}{c|}{Z (Acc)} & \multicolumn{1}{l}{}                          & \multicolumn{1}{l}{}                      & Z avg. & \multicolumn{1}{c}{Z (Acc)} & \multicolumn{1}{c}{Z (Acc)} & \multicolumn{1}{c}{Z (Acc)} \\ \hline
      {\color[HTML]{3531FF} high}                      & eng                   & 0.81   & 0.7 (53.6) & 1.33 (81.3) & 0.41 (75.6)                   & {\color[HTML]{3531FF} high}                                          & eng                                       & -0.56  & -0.49 (25) & -0.63 (44.2) & -0.58 (51.8)                  \\
      {\color[HTML]{3531FF} high}                      & \texttt{zho}                   & 0.79   & 1.11 (63.6) & 0.94 (73.9) & 0.31 (73.2)                  & {\color[HTML]{3531FF} high}                                          & \texttt{spa}                                       & -0.62  & -0.84 (16.4) & -0.87 (39.6) & -0.15 (62.2)                \\
\bottomrule
    \end{tabular}
  }

  \caption{Results for each task on \tool across 30 languages, evaluated using llama-3.3-70b-instruct, llama-3.1-8b-instruct, qwen2.5-72b-instruct and qwen2.5-7b-instruct. Accuracy was normalized using Z-scores and averaged across tasks. Languages were then ranked by their averaged Z-score.}
  \label{tab:all_result2}
\end{table*}

\end{document}